\documentclass[default,iicol]{sn-jnl}

\usepackage{tabularx}
\usepackage{array}

\usepackage{booktabs}
\usepackage{array}
\newcommand{\PreserveBackslash}[1]{\let\temp=\\#1\let\\=\temp}
\newcolumntype{C}[1]{>{\PreserveBackslash\centering}p{#1}}
\newcolumntype{R}[1]{>{\PreserveBackslash\raggedleft}p{#1}}
\newcolumntype{L}[1]{>{\PreserveBackslash\raggedright}p{#1}}
\usepackage{longtable}
\usepackage{xurl}

\jyear{2026}%

\theoremstyle{thmstyleone}%
\theoremstyle{thmstyletwo}%

\theoremstyle{thmstylethree}%

\usepackage[T1]{fontenc}

\begin{document}

\title[MOT Review]{In Pursuit of Many: A Review of Modern Multiple Object Tracking Systems}


\author*[1]{\fnm{Mk} \sur{Bashar}}\email{basharmk@msu.edu}
\author[1]{\fnm{Samia} \sur{Islam}}\email{islamsa3@msu.edu}
\author[2]{\fnm{Kashifa Kawaakib} \sur{Hussain}}\email{kashifa@iut-dhaka.edu}
\author[2]{\fnm{Md. Bakhtiar} \sur{Hasan}}\email{bakhtiarhasan@iut-dhaka.edu}
\author[2,3]{\fnm{A.B.M. Ashikur} \sur{Rahman}}\email{ashikiut@iut-dhaka.edu}
\author[2]{\fnm{Md. Hasanul} \sur{Kabir}}\email{hasanul@iut-dhaka.edu}

\affil[1]{\orgdiv{Department of Computer Science and Engineering}, \orgname{Michigan State University}, \orgaddress{\city{East Lansing}, \state{MI}, \country{USA}}}
\affil[2]{\orgdiv{Department of Computer Science and Engineering}, \orgname{Islamic University of Technology}, \orgaddress{\city{Gazipur}, \postcode{1704}, \country{Bangladesh}}}
\affil[3]{\orgdiv{Department of Information \& Computer Science}, \orgname{King Fahd University of Petroleum and Minerals}, \orgaddress{\city{Dhahran}, \country{Saudi Arabia}}}


\abstract{Multiple Object Tracking (MOT) is a core capability in modern computer vision, essential to autonomous driving, surveillance, sports analytics, robotics, and biomedical imaging. Persistent identity assignment across frames remains challenging in real scenes because of occlusion, dense crowds, appearance ambiguity, scale variation, camera motion, and identity switching. In this survey we synthesize recent progress by organizing methods around the problems they target and the paradigms they adopt. We cover the historical progression from tracking-by-detection to hybrid and end-to-end designs, and we summarize major architectural directions including transformer-based trackers, generative/diffusion formulations, state-space predictors, Siamese and graph-based models, and the growing impact of foundation models for detection and representation. We review benchmark trends that motivate method design, documenting the shift from saturated pedestrian benchmarks to challenge-driven and domain-specific datasets and we analyze evaluation practice by comparing classic and newer motion- and safety-centric metrics. Finally, we connect algorithmic trends to practical deployment constraints and outline emerging directions, foundation-model integration, open-vocabulary and multimodal tracking, unified evaluation, and domain-adaptive methods, that we believe will shape MOT research and real-world adoption.}




\maketitle

\section{Introduction}\label{sec1}
Multiple Object Tracking (MOT) links frame-level perception to temporally coherent scene understanding by localizing objects each frame and assigning persistent identities so that object trajectories can be formed. This temporal identity is essential for many high-impact applications such as autonomous driving, video analytics for public safety, sports analytics, and behavioral ecology, where a single-frame detector is not sufficient. MOT covers tracking multiple instances (often of the same class) simultaneously; by contrast, Single Object Tracking (SOT) follows a single target. MOT therefore must handle inter-object interactions, frequent occlusion, visually similar targets, small objects, and identity switches.

Over the last decade deep learning has reshaped vision pipelines, and MOT has followed that trend \cite{bewley2016simple, milan2017online}. Classical MOT combined hand-crafted appearance models, motion priors, and combinatorial data association; modern approaches increasingly rely on learned representations, end-to-end architectures, and temporal modelling to improve robustness to the failure modes above. Deep networks enable richer appearance features, learned motion/interaction priors, and sequence-level reasoning, but they also raise new design choices (e.g., tracking-by-detection versus end-to-end models, transformer-based sequence models) and evaluation challenges that merit systematic synthesis.

Existing surveys have helped shape the field but leave room for a complementary, challenge-driven synthesis. Some reviews concentrate on deep-learning pipelines and benchmark-driven comparisons \cite{pal2021deep,park2021multiple,du2024exploring,agrawal2024systematic}, others focus narrowly on data association \cite{luo2021multiple,rakai2021data}, and recent work expands categorizations and discusses end-to-end and transformer paradigms \cite{du2024exploring,guan2025multi}. Still, most surveys remain organized around methodological splits and leaderboard perspectives, with limited explicit mapping between concrete failure modes (for example long-term occlusion, severe crowding, irregular motion, and domain shift) and the design choices that address them. Moreover, deployment-focused constraints (real-time operation, heterogeneous sensors, safety-critical evaluation) and newer evaluation protocols (motion-centric, open-world, variable frame-rate) are discussed only at a high level in many reviews \cite{du2024exploring}.

Our survey fills these gaps by (i) reviewing a broad and recent body of work beyond just deep-learning pipelines, (ii) organizing methods around the concrete challenges they target, (iii) summarizing benchmark datasets and classical and novel evaluation metrics, and (iv) highlighting application constraints and promising directions for future work. We emphasize practical relevance and aim the paper at both newcomers, who need intuitive motivation and clear definitions, and experienced researchers, who benefit from a challenge-to-solution mapping and up-to-date metric and dataset analysis.

The main contributions are:
\begin{enumerate}
\item We present a concise, comprehensive overview of MOT and its role within modern vision systems.
\item We synthesize recent techniques from a study of over a hundred papers, categorizing methods according to the challenges they address and tracing recent trends.
\item We illustrate the practical relevance of MOT with diverse application examples and discuss deployment constraints.
\item We summarize popular benchmark datasets and evaluation metrics and introduce a discussion of novel metrics for more nuanced assessment.
\item We provide actionable insights and identify promising directions for future research in MOT.
\end{enumerate}

The remainder of the paper is organized as follows: In \autoref{sec:challenges} we examine the principal challenges in MOT and common failure modes. \autoref{sec:approaches} compiles the principal approaches used to address these challenges. \autoref{sec:datasets} surveys widely used benchmark datasets. \autoref{sec:metrics} reviews evaluation protocols and metrics. \autoref{sec:applications} discusses applications and deployment concerns. Finally, \autoref{sec:future} outlines directions for future work.

\section{Challenges in Multiple Object Tracking}
\label{sec:challenges}
Multiple Object Tracking is difficult because it must recover persistent identities from noisy, partial, and dynamic observations. The difficulties are not independent: poor detections amplify association errors, occlusion exacerbates identity switches, and computational constraints limit the sophistication of temporal models. Below we present the principal, interlinked challenge themes. For each theme we define the problem, explain why it arises in MOT settings, describe how it manifests empirically (for example as fragmentation, identity switches, or drift), summarize broad mitigation strategies, and note why the issue remains only partially solved.
\subsection{Occlusion}
\begin{figure*}[htb]
    \centering
    \includegraphics[width=.95\textwidth]{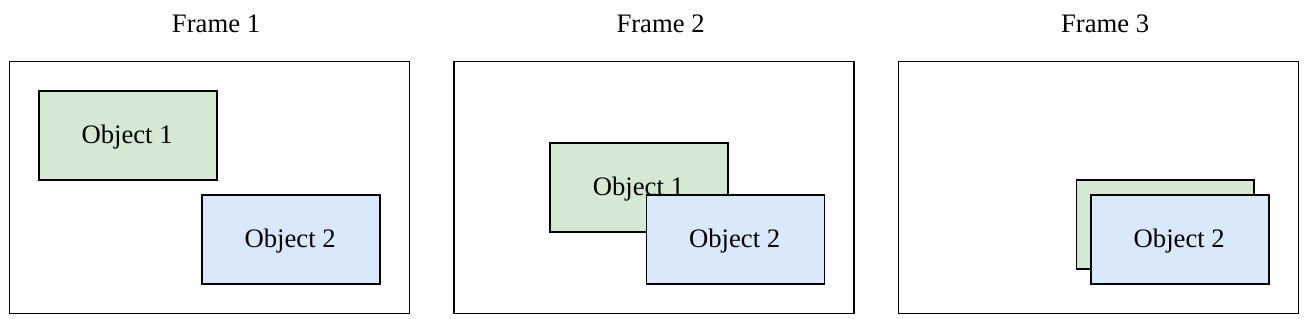}
    \caption{{\bf Illustration of the occlusion of two objects.} In frame 1, the two objects are separate from each other. In frame 2, the objects partially occlude each other. In frame 3, object 2 fully occludes object 1.}
    \label{fig:occlusionIllustration}
\end{figure*}

Occlusion occurs when targets are partially or fully hidden by other objects or scene elements. It arises from crowding, perspective, or camera viewpoint changes and is especially frequent in pedestrian and traffic scenes. As shown in \autoref{fig:occlusionIllustration}, multiple objects can be in occluded and non-occluded conditions in consecutive frames, posing a significant challenge for object tracking.

In practice occlusion causes detection dropouts, track fragmentation, spurious re-initializations, and identity switches; overlapping ground-truth bounding boxes further complicate training and evaluation in crowded scenes \cite{dendorfer2020mot20}.


Methods cope by (i) strengthening temporal context so tracks persist through missing detections (memory-augmented transformers, long short-term modules) \cite{gao2023memotr}, (ii) jointly modelling segmentation and tracking to resolve overlaps \cite{meinhardt2021trackformer}, and (iii) using global graph or appearance models to reason across frames \cite{huo2021multi,milan2017online,tian2018detection,ullah2018directed}. Recent practical advances also focus on post-detection processing that preserves low-score but informative detections (occlusion-aware NMS) and motion-centric association to reduce ID switches in crowded, non-linear scenes \cite{huang2024deconfusetrack,cao2023observation}.
Despite progress, long-term occlusions and severe crowding remain open because they require reliable long-horizon identity priors, robust re-identification, or richer scene understanding without incurring prohibitive complexity.

\subsection{Detection Noise and Background Complexity}
Detection noise denotes false positives, missed detections, and score instability caused by cluttered backgrounds, dynamic scene elements, camera motion, or low-quality imaging. These phenomena make it hard to separate foreground from background and produce false tracks, track drift, and interrupted trajectories. Camera motion and moving background elements induce spurious motion cues and blur, while similar textures and dense scenes reduce detector confidence.

Remedies include background modelling and motion compensation, detector architectures with attention or contextual reasoning, and keeping more candidates for downstream association rather than aggressive filtering \cite{zhang2022bytetrack}. End-to-end tracking systems often underperform due to weak detection components; bootstrapping from stronger detectors or hybrid training strategies can help, as in MOTRv2 and candidate-preserving association designs \cite{zhang2023motrv2,zhang2022bytetrack}.
The challenge endures because detector robustness depends on data diversity, domain gap mitigation, and the trade-off between pruning false alarms and retaining occluded or low-score true positives.

\subsection{Detection Speed and Real-time Constraints}
Detection latency limits the usable frame rate and directly affects tracking continuity in real-time systems. Large, high-capacity detectors and unoptimized implementations increase inference time, causing missed frames, delayed associations, and degraded tracking in time-sensitive applications. Approaches to mitigate this trade-off include architectural efficiency (lightweight backbones, deformable attention), system-level optimizations, model compression, and hardware acceleration \cite{yan2021lighttrack,zhu2021vitt,yu2022relationtrack,blatter2021efficient,zhu2021vitt}.

Recent work also explores training and label-assignment strategies that narrow the performance gap between end-to-end and tracking-by-detection while reducing compute \cite{yang2024hybrid}. The tension between throughput and accuracy persists, especially for high-resolution inputs and dense scenes.

\subsection{Lighting and Imaging Degradation}
Variations in illumination, shadows, nighttime conditions, and sensor noise alter appearance and reduce detector reliability, producing missed detections and identity confusion. Solutions range from preprocessing (illumination normalization, de-noising), multimodal sensing (infrared, depth), to dedicated low-light training and modules that explicitly model degradation \cite{wang2024multi}.

Robust feature learning mitigates some issues, but domain-specific degradations and generalization to unseen lighting conditions remain challenging without targeted data or domain adaptation.


\begin{figure*}[htb]
    \centering
    \includegraphics[width=.95\textwidth]{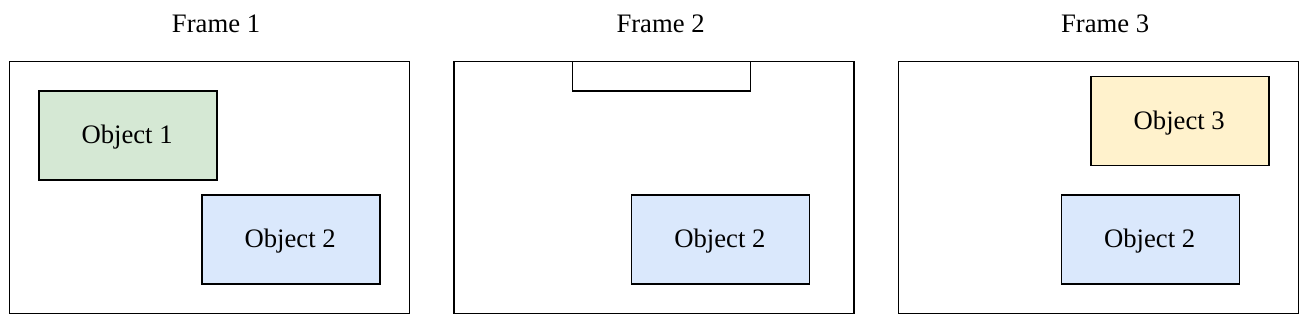}
    \caption{{\bf Illustration of the ID Switching of an object.} In frame 1, Object 1 is detected. In frame 2, Object 1 goes slightly out of frame and hence becomes not detected. In frame 3, Object 1 re-enters the frame but is detected as a new Object 3.}
    \label{fig:idswitchIllustration}
\end{figure*}


\subsection{Identity Switching}
Identity switching refers to erroneous reassignment of an existing identity to a different object, or creation of duplicate identities for the same object (\autoref{fig:idswitchIllustration}). It is triggered by similar appearances, abrupt motion, occlusions, or detector instability and results in fragmented trajectories and incorrect counts. Common remedies include memory and re-identification buffers, multi-frame/global association algorithms, graph-based optimization, and verification/correction mechanisms \cite{cai2022memot,zhou2022global,dai2021learning}. Techniques that decompose association under ambiguity and preserve occluded detections have shown empirical gains \cite{huang2024deconfusetrack,gao2023memotr}. Nonetheless, maintaining consistent identity over long gaps and across severe appearance changes remains only partially addressed.

\subsection{Scale Variation}
Objects vary widely in scale due to perspective and distance, producing missed small-object detections and ambiguous matches across frames. Scale-aware detectors, multi-scale feature hierarchies, and dynamic scale estimation based on motion and context alleviate these failures \cite{lin2021swintrack}.

Data augmentation and hierarchical frameworks improve robustness, but extreme scale ranges in real scenes still challenge detector resolution and association reliability.


\subsection{Model Size and Deployment Constraints}
Practical MOT requires lightweight architectures for low-latency processing on edge devices. Heavy-weight models improve accuracy but are impractical for real-time deployment in robotics, surveillance, or autonomous vehicles. Strategies include compact backbones, knowledge distillation, and algorithmic simplifications (e.g. sparse attention) with acceptable accuracy trade-offs \cite{zhu2021vitt,yu2022relationtrack,yan2021lighttrack}. Balancing model capacity, pretraining benefits, and runtime constraints remains an active engineering and research problem.


\begin{figure*}[htb]
    \centering
    \includegraphics[width=.8\textwidth]{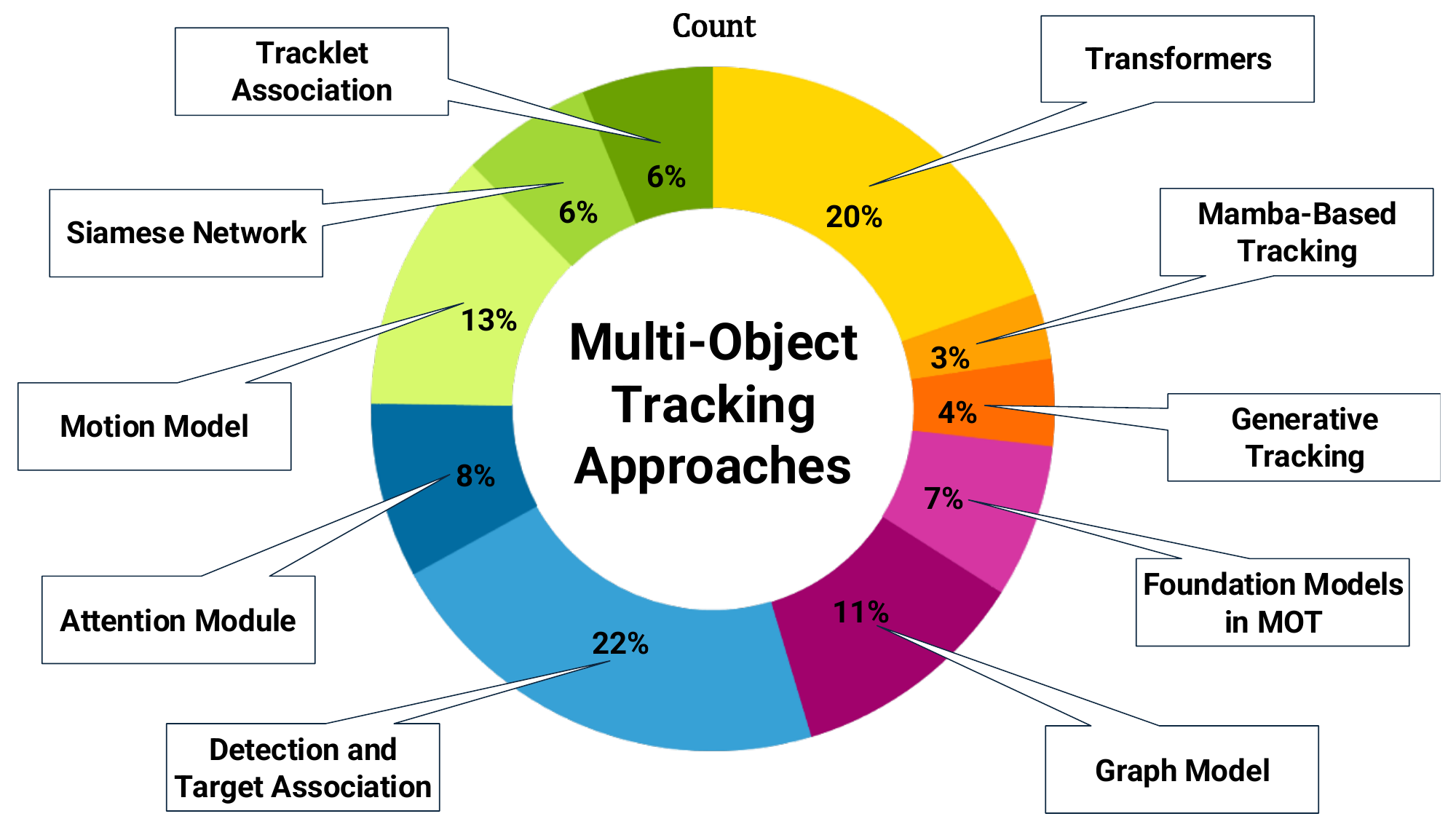}
    \caption{\bf Piechart based on the paper count of each approaches for multi-object tracking.}
    \label{fig:approaches}
\end{figure*}

\subsection{Miscellaneous}
Other persistent issues include motion blur from camera or object motion, which degrades appearance features; robustly determining when to start or terminate tracks to avoid spurious short-lived tracklets; and non-rigid deformations or inter-class appearance similarity that impede reliable discrimination \cite{chong2021overview}. These problems are addressed by specialized deblurring, temporal consistency checks, and leveraging contextual cues, but they persist because they often require per-domain solutions or richer scene semantics.



\section{Approaches of Tracking Multiple Objects}\label{sec:approaches}
We organize methods according to the dominant modeling choice and the assumptions it brings, and we present groups roughly in descending order of community activity (\autoref{fig:approaches}). This taxonomy is intended to help readers pick approaches by design principle (what is assumed about detections, temporal signal, and computation) rather than by implementation details. For each group we (i) state the unifying idea and core assumptions, (ii) describe the canonical pipeline and the role each module plays, (iii) highlight representative variations and how they address core MOT failure modes, and (iv) critically reflect on strengths, limitations, and when one should prefer alternatives.

\subsection{Detection and Target Association}
\label{subsection:detection_association_mot}
Tracking-by-detection assumes a per-frame detector supplies candidate boxes or masks and that appearance plus short-term motion suffice to resolve correspondence in most frames. The core assumption is that detection failures are either rare or can be mitigated by preserving low-score candidates for association.

\begin{table*}
\centering
\caption{Summary of Detection and Target Association based approaches}
\label{tab:detection_association}
\footnotesize

\begin{tabular}{
  C{0.03\textwidth}
  C{0.04\textwidth}L{0.18\textwidth}
  L{0.23\textwidth}
  L{0.23\textwidth}
  L{0.12\textwidth}
}
\toprule
\textbf{Ref.} & \textbf{Year} & \textbf{Detection} & \textbf{Association} & \textbf{Dataset} & \textbf{MOTA (\%)} \\
\midrule

\cite{keuper2018motion} & 2018 &
Faster R-CNN &
Correlation Co-Clustering &
MOT15, MOT16, MOT17 &
35.6, 47.1, 51.2 \\

\cite{karunasekera2019multiple} & 2019 &
DPM, F-RCNN, SDP, RRC &
Hungarian Algorithm &
MOT17, KITTI &
46.9, 85.04 \\

\cite{voigtlaender2019mots} & 2019 &
Mask R-CNN &
Distance Measurement &
KITTI MOTS, MOTSChallenge &
65.1, 66.9 (MOTSA) \\

\cite{zhang2021fairmot} & 2021 &
CenterNet (DLA-34) / JDE &
ReID embedding + Kalman Filter + Hungarian Algorithm &
MOT15, MOT16, MOT17, MOT20 &
60.6, 74.9, 73.7, 61.8 \\

\cite{muller2021seeing} & 2021 &
Sparse 3D CNN + Multi-head 3D object detector &
Hungarian Algorithm &
DYNSYNTH &
42.3 \\

\cite{song2021multiple} & 2021 &
CenterNet &
Hungarian Algorithm &
MOT15, MOT16, MOT17, MOT20 &
60.6, 74.9, 73.7, 61.8 \\

\cite{wang2021drt} & 2021 &
ResNet50 &
LSTM-based Motion Model &
MOT16, MOT17 &
76.3, 76.4 \\

\cite{kim2021discriminative} & 2021 &
CenterNet &
Bilinear LSTM &
MOT16, MOT17 &
48.3, 51.5 \\

\cite{wang2021multiple} & 2021 &
CenterNet &
Correlation Learning &
MOT15, MOT16, MOT17, MOT20 &
62.3, 76.6, 76.5, 65.2 \\

\cite{pang2021quasi} & 2021 &
Faster R-CNN &
Quasi-dense Similarity Matching &
MOT16, MOT17, BDD100K, Waymo &
69.8, 68.7, 64.3, 51.18 \\

\cite{sundararaman2021tracking} & 2021 &
HeadHunter &
HeadHunter-T &
CroHD &
63.6 \\

\cite{wu2021track} & 2021 &
CenterNet &
CVA (Cost Volume based Association) &
MOT16, MOT17, nuScenes, MOTS &
70.1, 69.1, 5.9 (AMOTA), 65.5 (MOTSA) \\

\cite{liu2022det} & 2022 &
Mask-RCNN &
Hungarian Algorithm &
MOT17, MOT20, NTU-MOTD &
43.21, 57.70, 92.12 \\

\cite{tan2022towards} & 2022 &
YOLOv4 &
Hungarian Algorithm &
TAMU2015V, UGA2015V, UGA2018V &
79.0, 65.5, 73.4 \\

\cite{kesa2022multiple} & 2022 &
DLA-34 &
Hungarian Algorithm &
MOT15, MOT16, MOT17, MOT20 &
55.8, 73.8, 74.0, 60.2 \\

\cite{sun2022online} & 2022 &
DPM and YOLOv5 with detection modifier (DM) &
Global and Partial Feature Matching &
MOT16 &
46.5 \\

\cite{liang2022non} & 2022 &
YOLO X with later NMS &
Kalman Filtering, Bicubic Interpolation and ReID Model &
MOT17, MOT20 &
78.3, 75.7 \\

\cite{he2022joint} & 2022 &
T-ReDet module &
ReID-NMS Model &
MOT16, MOT17, MOT20 &
63.9, 62.5, 57.4 \\

\cite{yang2024hybrid} & 2024 &
YOLOX &
Hybrid-SORT + Hungarian Algorithm &
MOT17, MOT20, DanceTrack &
79.3, 76.4, 91.6 \\

\cite{yi2024ucmctrack} & 2024 &
Public detections (e.g., YOLOX/ByteTrack detections) &
Uniform Camera Motion Compensation + ground-plane motion association &
MOT17, MOT20, DanceTrack &
79.0, 75.5, 88.8 \\

\cite{kim2025plugtrack} & 2025 &
YOLOX &
Adaptive fusion of Kalman + data-driven motion (PlugTrack) &
MOT17, MOT20, DanceTrack &
79.2, 76.4, 92.4 \\

\bottomrule
\end{tabular}
\end{table*}

The canonical pipeline for detection and target association-based MOT approaches contains four conceptual blocks: (1) a detector producing boxes or masks and confidence scores, (2) an embedding head that yields per-candidate appearance descriptors, (3) a prediction module (for example, Kalman filter or learned predictor) that extrapolates track states, and (4) an association module that computes affinities (IoU, appearance cosine, motion cost) and solves matching (greedy, Hungarian, or learned matching). Post-processing implements track initiation, termination, and re-identification.

Several strands of work refine the detector, the embedding head, the prediction module, or the association cost and matching solver to improve robustness and efficiency. A summary of the works in detection and target association based approaches can be found in \autoref{tab:detection_association}.

Keuper et al. combined point trajectories with bounding-box reasoning to strengthen short-term linking and motion cues \cite{keuper2018motion}. Voigtlaender et al. extended detection to segmentation with Track-RCNN to provide mask-level cues for overlap resolution \cite{voigtlaender2019mots}. Liang et al. introduced Non-Maskable Suppression and He et al. proposed joint re-detection and re-identification trackers that explicitly focus on bounding-box preservation and robust re-detection under heavy occlusion \cite{liang2022non, he2022joint}.

A second family integrates embedding extraction into detection. FairMOT balances detection and re-identification in a single-stage dual-branch design to avoid mismatched objectives between separate detector and re-ID networks \cite{zhang2021fairmot}. QDTrack fuses two-stage detectors such as Faster-RCNN with similarity learning to improve discriminativeness in crowded scenes \cite{pang2021quasi}. Sundararaman et al. designed HeadHunter and HeadHunter-T to combine advanced detection backbones (FPN, ResNet-50, RPN) with tracking heads for small and crowded object detection \cite{sundararaman2021tracking}. Sun et al. improved feature extraction that benefits both detection and association \cite{sun2022online}.

Occlusion and candidate preservation have motivated practical heuristics. Detection Refinement for Tracking (DRT) uses semi-supervised learning and LSTM modules to keep detections persistent through occlusions \cite{wang2021drt}. ByteTrack and related pipelines deliberately preserve low-score detections so downstream matching can recover true positives lost due to occlusion or blur \cite{zhang2022bytetrack}. DeconfuseTrack proposes occlusion-aware non-maximum suppression and decomposed association to reduce ID switches under ambiguity \cite{huang2024deconfusetrack}.

Motion and geometry priors complement appearance learning. Hybrid-SORT incorporates confidence and height cues into association costs to handle non-linear motion \cite{yang2024hybrid}. Geometry-centric trackers and camera-motion compensation (UCMCtrack variants) show that careful geometric priors remain competitive when learned re-ID fails \cite{yi2024ucmctrack}. CorrTracker and correlational network designs emphasize correlation-based affinity modeling for robustness in dense interactions \cite{wang2021multiple}. Application-driven works adapt detection and association to domain dynamics such as indoor scenes, agriculture, forecasting, and sports \cite{liu2022det, tan2022towards, kesa2022multiple}. Finally, hybrid training and lightweight encoders narrow the gap between end-to-end and detect-then-associate approaches while reducing compute \cite{yang2024hybrid, zhang2023motrv2}.

The paradigm is modular, efficient, and interpretable for deployment. Its main failure modes are detector-driven: missed or unstable detections cause fragmentation and ID switches, while aggressive pruning trades away recall. Modular pipelines require hand-tuned heuristics for thresholds and birth/death policies. Relative to end-to-end transformers, detection+association is typically less data-hungry and more practical for real-time systems but weaker at long-horizon global reasoning.


\subsection{Transformers}
\label{subsection:transformer}
Transformer-based methods use attention to learn global spatial and temporal interactions. They assume that a set of queries, partitioned into detection and track queries, can represent persistent entities and that cross-attention to feature maps can both localize objects and resolve correspondence.

\begin{table*}
\caption{Summary of Transformer based approaches}
\label{tab:transformers}
\footnotesize
    \begin{tabular}
    {C{0.03\textwidth}
  C{0.04\textwidth}L{0.18\textwidth}
  L{0.23\textwidth}
  L{0.23\textwidth}
  L{0.12\textwidth}}
    \toprule
    \textbf{Ref.} & \textbf{Year} & \textbf{Detection/Appearance Feature Extraction} & \textbf{Data Association} & \textbf{Dataset(s)} & \textbf{MOTA (\%)} \\
    \midrule
    \cite{sun2020transtrack} & 2020 & Decoder of DETR & Decoder of Transformer & MOT17, MOT20 & 74.5, 64.5 \\

    \cite{meinhardt2021trackformer} & 2021 & CNN & Decoder of Transformer & MOT17 & 62.5 \\
    \cite{xu2021transcenter} & 2021 & DETR & Deformable Dual Decoder & MOT17, MOT20 & 71.9, 62.3 \\
    \cite{zeng2021motr} & 2021 & DETR & Decoder + Query Interaction Module + Temporal Aggregation Network & MOT17, DanceTrack, BDD100k & 57.2 (HOTA), 54.2 (HOTA), 32.0 (nMOTA) \\
    \cite{zhu2021vitt} & 2021 & Encoder & Bounding Box Regression Network & MOT16 & 65.7 \\
    \cite{blatter2021efficient} & 2021 & Exemplar Attention based encoder & Exemplar Attention based encoder & TrackingNet & 70.55 (Precision) \\

    \cite{chen2022patchtrack} & 2022 & CNN (patch extraction guided by motion) & Transformer & MOT16, MOT17 & 73.3, 73.6 \\
    \cite{liu2022segdq} & 2022 & CNN + Encoder of Transformer & Decoder + Feed Forward Network & MOT15, MOT16, MOT17 & 40.3, 65.7, 65.0 \\
    \cite{xing2022siamese} & 2022 & Transformer Pyramid Network & Multihead and pooling attention & UAV123 & 85.83 (Precision) \\
    \cite{zhou2022global} & 2022 & CenterNet & Tracking transformer & TAO, MOT17 & 45.8 (HOTA), 75.3 \\
    \cite{yu2022relationtrack} & 2022 & Faster R-CNN & Hungarian Algorithm & MOT16, MOT17, MOT20 & 75.8, 74.7, 70.5 \\
    \cite{cai2022memot} & 2022 & Transformer-based Network & Memory Encoding and Decoding & MOT16, MOT17, MOT20 & 72.6, 72.5, 63.7 \\
    \cite{ma2022unified} & 2022 & Detection Results of FairMOT & Track Transformer & MOT16 & 74.2 \\
    \cite{zhang2023motrv2} & 2023 & ResNet-50 backbone + YOLOX proposal queries (bootstrapped detector) + Deformable encoder/decoder & Query-based tracking via propagated track queries (MOTR-style) & MOT17, MOT20, DanceTrack & 78.6, 76.2, 91.9 \\
    \cite{yu2023motrv3} & 2023 & Track query groups + auxiliary boxes from pretrained detectors (e.g., YOLOX/Sparse R-CNN) & Track query group update + Release-Fetch supervision for association robustness & MOT17, DanceTrack & 75.9, 92.9 \\

    \multirow[t]{5}{*}{\cite{luo2025co}} & \multirow[t]{5}{*}{2025} & CNN backbone + Deformable encoder/decoder (MOTR-style) + Shadow Sets & Coopetition Label Assignment + Shadow queries for newborn/track competition & MOT17, MOT20, DanceTrack & 72.6, 60.1, 89.3 \\
    \multirow[t]{3}{*}{\cite{zeng2025tgformer}} & \multirow[t]{3}{*}{2025} & DAB-Deformable-DETR detector (COCO pretrained) & Track Query Group mechanism (multiple queries per target) & MOT17, MOT20, DanceTrack (private det.) & \multirow[t]{3}{*}{74.9, 70.3, 91.3} \\
    \multirow[t]{3}{*}{\cite{gao2025multiple}} & \multirow[t]{3}{*}{2025} & Deformable DETR features + ID tokens (identity prompts) & In-context ID prediction via ID Decoder (association as classification) & DanceTrack, SportsMOT & \multirow[t]{3}{*}{90.6, 92.4} \\
    \multirow[t]{3}{*}{\cite{liao2024fasttracktr}} & \multirow[t]{3}{*}{2025} & RT-DETR detector + ID Embedding head (JDT-style) & Similarity matrix on embeddings + Hungarian + Kalman smoothing & \multirow[t]{3}{*}{MOT17, DanceTrack} & \multirow[t]{3}{*}{76.7, 88.8} \\
    \bottomrule
    \end{tabular}

\end{table*}

In this method, typically a backbone (CNN or ViT) produces features; a transformer encoder builds contextual maps; a transformer decoder operates on queries split into detection queries and track queries. Detection queries find new objects; track queries are propagated or updated to maintain identities. Association is realized implicitly when decoder outputs indicate whether a track query matches a current observation. Variants add memory, multiple queries per target, or query-selection heuristics to manage occlusion and newborn objects. 

Transformer trackers split into several motifs: pure query-propagation end-to-end models, bootstrapped hybrids, efficiency-focused sparse-attention variants, memory-augmented decoders, and designs that decouple detection from association. See \autoref{tab:transformers} for a comparative view.


Early query formulations cast detection and tracking as unified matching problems. Xu et al. (TransCenter), Transtrack, and TrackFormer used detection and track queries propagated across frames \cite{xu2021transcenter, sun2020transtrack, meinhardt2021trackformer}. Ma et al. proposed a Unified Transformer Tracker demonstrating that a single transformer trained with unified objectives can generalize across single-object and multi-object settings \cite{ma2022unified}.

Optimization and supervision challenges spurred hybrids. MOTRv2 bootstraps decoder queries from a strong external detector to reduce the decoder learning burden \cite{zhang2023motrv2}. MOTRv3 introduces release–fetch supervision to rebalance detection and tracking losses \cite{yu2023motrv3}. CO-MOT uses coopetition-based label assignment by expanding each query into auxiliary shadow queries, which increases positive supervision for newborn objects during training \cite{luo2025co}. These methods address the sparse-positive-sample and slow-convergence issues in early transformer trackers.

Appearance and motion trade-offs produced mixed strategies. Chen et al. suggest offloading short-range motion prediction to light motion models and letting the transformer focus on appearance matching, which reduces the transformer's burden \cite{chen2022patchtrack}. Yu et al. proposed Global Context Disentanglement to separate representations for detection and for association inside one-stage transformers to reduce conflicting gradients \cite{yu2022relationtrack}.

Memory and redundancy improve long-term persistence. Cai et al. (MeMOT) developed explicit short- and long-term memory banks combined with cross-attention to stabilize track embeddings across occlusions \cite{cai2022memot}. Zhou et al. processed multiple frames concurrently via cross-attention for longer-term association \cite{zhou2022global}. TGFormer and related multi-query-per-target designs assign multiple queries per object so different queries can specialize for distinct visibility states such as fully visible or heavily occluded \cite{zeng2025tgformer}.

Efficiency-focused variants address the computational burden. Zhu et al. used lightweight encoders and feed-forward tracking heads \cite{zhu2021vitt}. Blatter et al. proposed Exemplar Attention leveraging shared memory to reduce per-object cost \cite{blatter2021efficient}. Zeng et al. used Query Interaction Modules and Tracklet-Aware Label Assignment to filter irrelevant queries and enforce one-object one-query mapping \cite{zeng2021motr}. FastTrackTr rethinks cross-decoder flows to implicitly incorporate historical information without explicit track queries, improving speed \cite{liao2024fasttracktr}. MOTIP reframes association as identity prediction using an ID decoder, eliminating explicit cost matrices \cite{gao2025multiple}.

Transformers excel at global reasoning and can learn association end-to-end, reducing handcrafted heuristics. Their limits are practical: high compute and data demands, training instability when objectives compete, and poor recall for newborn objects unless special measures are taken. Hybrid and memory-aware designs mitigate some issues, making transformers increasingly practical.

\subsection{Motion Model}
Motion-based approaches assume object dynamics and scene geometry provide reliable priors for association, especially when appearance is degraded. Typically they assume locally smooth dynamics or that dynamics can be learned with sufficient data.

\begin{table*}
\centering
\caption{Summary of Motion Model based approaches}
	\label{tabmotion}
    \footnotesize
    \begin{tabular}
    {C{0.03\textwidth}
  C{0.04\textwidth}
  L{0.41\textwidth}
  L{0.23\textwidth}
  L{0.12\textwidth}}
    \toprule
	\textbf{Ref.} & \textbf{Year} & \textbf{Motion Mechanism} & \textbf{Dataset(s)} & \textbf{MOTA (\%)} \\
	\hline
	\cite{karunasekera2019multiple} &  2019 & Dissimilarity Distance between Detected and Predicted Object & MOT17, KITTI &  46.9, 85.04 \\
\cite{kesa2021joint} & 2021 & Dissimilarity Distance between Detected and Predicted object & MOT15, MOT16, MOT17, MOT20 & 55.8, 73.8, 74.0, 60.2 \\
\cite{wang2021drt} & 2021 & LSTM-based Model on Consecutive Frames &  MOT16, MOT17 &  76.3, 76.4 \\
\cite{qin2021joint} & 2021 & Kalman Filtering & MOT17 &  44.3 \\
\cite{yin2021detecting} & 2021 & Accumulative Multi-Frame Differencing and Low-Rank Matrix Completion &  VISO &  73.6 \\
\cite{shi2021anomalous} & 2021 & Distance of Motion Feature and Mean Vector of Gaussian Local Velocity Model &  NJDOT & 100 (Anomaly Detection Accuracy) \\
\cite{wang2021track} & 2021 & Box and Tracklet Motion Embedding & MOT17, KITTI, UA-Detrac &56.0, 87.6, 22.5 \\
\cite{sundararaman2021tracking}   &  2021 & Particle Filtering and Enhanced Correlation Coefficient Maximization &  CroHD &  63.6 \\
\cite{han2022mat} &  2022 & Combination of Camera Motion and Pedestrian Motion (IML), Dynamic Motion-based Reconnection (DRC) &  MOT16, MOT17 &  70.5, 69.5 \\
\cite{zou2022compensation}  &  2022 & Motion Compensation with Basic Tracker & MOT16, MOT17, MOT20 & 69.8, 68.8. 66.0 \\
\cite{chen2022patchtrack}  &  2022 &  Kalman Filtering & MOT16, MOT17 & 73.3, 73.6 \\
\cite{liu2022multi} & 2022 &  Adaptive Motion Filter (AMF) & VisDrone2019, UAVDT & 36.1, 46.4\\
\bottomrule
	\end{tabular}

\end{table*}

In this method, typically a motion estimator predicts future track locations (Kalman, linear predictors, learned sequence models). Predictions prune unlikely matches and contribute to motion-consistency costs. Motion can be fused with appearance in affinity networks or used to guide patch extraction for re-id. See \autoref{tabmotion} for a summary of the works in motion-based approaches.

Classical motion priors such as Kalman filters and constant-velocity models remain common baselines. Karunasekera et al. used predicted-versus-observed discrepancies to inform association costs \cite{karunasekera2019multiple}. Learned recurrent predictors address occlusion and non-linear motion; Wang et al. employed LSTM-based motion models to extrapolate through occlusions \cite{wang2021drt}. Qin et al. combined motion prediction with Deep Affinity Networks to constrain association regions \cite{qin2021joint}. Han et al. (MAT) jointly encoded motion cues for prediction and association \cite{han2022mat}.

Motion also aids detection and appearance extraction. Yin et al. used accumulative multi-frame differencing and low-rank completion to form a Motion Model Baseline for satellite imagery detection \cite{yin2021detecting}. Shi et al. leveraged motion features versus a global-local variance model for autonomous driving \cite{shi2021anomalous}. Zou et al. introduced motion compensation to recover objects lost to camera egomotion \cite{zou2022compensation}. Chen et al. used motion prediction to crop patches for re-ID, reducing background contamination \cite{chen2022patchtrack}. For UAVs and irregular motion, Liu et al. proposed local motion models with adaptive motion filters \cite{liu2022multi}. Modern predictors replace linear predictors with diffusion or SSM-based predictors to capture highly non-linear behaviors \cite{lv2024diffmot}.

Motion priors are computationally cheap and effective in structured scenes. They fail under abrupt maneuvers, extreme camera motion, or highly nonstationary dynamics. Learned motion models and compensation strategies reduce these failure modes, and motion modules integrate naturally with other families.

\subsection{Graph Model}
Graph models represent detections or tracklets as graph nodes and cast association as edge prediction or global optimization. They assume pairwise and higher-order affinities, propagated by message passing, can resolve ambiguities that local matching cannot.


\begin{table*}
    \centering
    \caption{Summary of Graph Model based approaches}
    \label{tabgcn}
    \footnotesize
    \begin{tabular}
    {C{0.03\textwidth}
  C{0.04\textwidth}L{0.18\textwidth}
  L{0.23\textwidth}
  L{0.23\textwidth}
  L{0.12\textwidth}}
    \toprule
    \textbf{Ref.} & \textbf{Year} & \textbf{Detection} & \textbf{Association} & \textbf{Dataset(s)} & \textbf{MOTA (\%)} \\
    \midrule
    \cite{braso2020learning} & 2020 & ResNet50 & Message Passing & MOT15, MOT16, MOT17 & 51.5, 58.6, 58.8 \\
\cite{li2020graph}  & 2020 & ResNet-34 & Hungarian algorithm & MOT16, MOT17 & 47.7, 50.2 \\

\cite{ma2021deep} & 2021 & SeResNet-50 & Human-Interaction Model & MOT15, MOT16, DukeMTMCT & 80.4, 50.0, 86.7 \\
\cite{wang2021track} & 2021 & CenterNet, CompACT & Box and Tracklet Motion Embedding & MOT17, KITTI, UA-Detrac & 56.0, 87.6, 22.5 \\
\cite{dai2021learning} & 2021 & ResNet50-IB & Proposal Generation and Scoring & MOT17, MOT20 & 59.0, 56.3 \\
\cite{he2021learnable} & 2021 & CenterNet & Graph Matching & MOT16, MOT17 & 65.0, 66.2 \\
\cite{quach2021dyglip} & 2021 & Per-camera tracker (e.g., DeepSORT tracklets + ReID) & Dynamic graph link prediction (DyGLIP) & PETS09, CAMPUS, EPFL, MCT, CityFlow & 93.5, 72.8, 66.3, 95.7, 90.9 \\
\cite{zaech2022learnable} & 2022 & CenterPoint, MEGVII & Message Passing & nuScenes & 57.0, 88.6 \\
\cite{buchner20223d} & 2022 & CenterPoint/PointRCNN & Cross-edge attention + message passing + trajectory clustering & nuScenes, KITTI & 76.7 (3D MOTA)\\
\cite{wang2024gslamot} & 2024 & PointPillars / MPPNet & Tracklet \& Query Graph + MSGA + OGO & Waymo Open Dataset, TCD & 59.69 (WOD), 49.10 (TCD) \\
\cite{zhang2025multiple} & 2025 & YOLOX & Weighted-GCNN association (learned edge weights) & MOT16, MOT17, MOT20 & 79.89, 80.27, 77.63 \\
    \bottomrule
    \end{tabular}

\end{table*}

The graph models generally build a graph (nodes are detections or tracklets), compute pairwise features (appearance, motion, geometry), apply a GNN or message passing to refine node and edge embeddings, then decode edges into associations followed by a global assignment or ranking stage. Tracklet-level graphs raise the temporal abstraction and reduce graph size. See \autoref{tabgcn} for a summary of graph-based approaches in MOT.

Graph methods differ by node granularity, dual-graph decompositions, and dynamic edge formulations. Braso and Leal-Taixé used message passing networks to learn global affinities across full sequences \cite{braso2020learning}. Zaech et al. and Ma et al. refined graph learning dynamics to better propagate appearance and motion cues \cite{zaech2022learnable, ma2021deep}. Li et al. proposed parallel Appearance and Motion Graph Networks to independently model visual similarity and kinematic consistency before fusion \cite{li2020graph}. Dai et al. used a two-stage graph with dense proposals that are pruned and re-scored by a GCN ranking module \cite{dai2021learning}.

Higher-order solvers and learned assignment layers improve global consistency. He et al. integrated quadratic-programming layers for globally consistent associations \cite{he2021learnable}. Tracklet-level graphs reason over longer horizons and better recover fragmented tracks \cite{wang2024gslamot}. Zhang et al. introduced dynamic weighted graphs with learnable edge confidences and iterative refinement to softly suppress unlikely associations instead of early hard pruning \cite{zhang2025multiple}.

Multi-camera and multimodal graphs connect nodes across views and sensors. Quach et al. built dynamic graphs that evolve with new observations for multi-camera accumulation \cite{quach2021dyglip}. For autonomous driving, modality-conditioned edges allocate attention to LiDAR when visual cues fail \cite{buchner20223d}. Graphs naturally fuse modalities for robust association across challenging conditions.

Graphs enable global, multi-frame reasoning and excel at recovering long-term tracks after fragmentation. They may be computationally heavy and sensitive to noisy node features and spurious edges. Scalable graph construction and effective pruning strategies are required for long videos. Graph approaches are complementary to transformers and tracklet methods.

\subsection{Attention Module}
Attention isolates discriminative foreground features and suppresses background clutter, and explicit memory stores per-track templates. The assumption is that attended regions and curated memory slots capture identity cues necessary to re-identify objects despite occlusion and appearance drift.


\begin{table*}
\centering
\caption{Summary of Attention based approaches}
	\label{tab:att}
    \footnotesize
    \begin{tabular}
    {C{0.03\textwidth}
  C{0.04\textwidth}
  L{0.36\textwidth}
  L{0.28\textwidth}
  L{0.12\textwidth}}
    \toprule
	\textbf{Ref.} & \textbf{Year} & \textbf{Attention Mechanism} & \textbf{Dataset(s)} & \textbf{MOTA (\%)} \\
	\midrule
	\cite{song2021multiple} & 2021 & Strip Pooling & MOT15, MOT16, MOT17, MOT20 & 60.6, 74.9, 73.7, 61.8 \\
\cite{guo2021online} & 2021 & Temporal Aware Target Attention and Distractor Attention & MOT16, MOT17, MOT20 & 59.1, 59.7, 56.6 \\
\cite{liang2021generic} & 2021 & Spatial Transformation  Network (STN) & MOT16, MOT17 & 50.5, 50.0 \\
\cite{ke2021prototypical} & 2021 & Spatio-Temporal Cross-Attention & BDD100K (Validation), KITTI-MOTS (Validation) & 27.4 (MOTSA), 66.4 (mMOTSA) \\
\cite{fu2021real} & 2021 & Self-Attention in Detection & Custom Dataset: Sparse Scene, Dense Scene & 70.9, 56.4 \\
\cite{quach2021dyglip} & 2021 & Graph Structural and Temporal Self-Attention & PETS09, EPFL, CAMPUS, MCT, CityFlow & 93.5, 66.3, 96.7, 95.7, 90.9 \\
\cite{wan2022dsrrtracker} & 2022 & Self- and Cross-Attention as Tracking Head & MOT17, MOT20 & 75.6, 70.4 \\
\cite{cai2022memot} & 2022 & Self- and Cross-Attention & MOT16, MOT17, MOT20 & 72.6, 72.5, 63.7 \\ 
	\bottomrule
	\end{tabular}

\end{table*}

The pipelines utilizing attention modules, insert spatial or channel attention into embedding heads to focus representation on foreground cues. They maintain per-track memory banks (short-term and long-term) that are queried using cross-attention to compute affinities for association and re-id. Here, memory update and selection policies determine which frames are retained. See \autoref{tab:att} for a summary of the recent works on MOT that utilize the attention module.

Attention modules range from spatial and strip attention to memory-augmented decoders. Song et al. used strip attention with combined max and mean pooling to emphasize pedestrian features under clutter \cite{song2021multiple}. Liang et al. applied Spatial Transformer Networks to constrain embeddings to foreground regions \cite{liang2021generic}. Ke et al. proposed Prototypical Cross-Attention and the PCAN to propagate discriminative foreground–background contrasts from history \cite{ke2021prototypical}. Guo et al. designed dual-attention frameworks that separate target and background feature processing and apply memory aggregation for consistency \cite{guo2021online}. Wan et al. combined self- and cross-attention in a lightweight architecture to balance accuracy and runtime \cite{wan2022dsrrtracker}.

Memory-augmented decoders such as MeMOT maintain short- and long-term memory banks and use cross-attention to stabilize track embeddings across extended occlusions \cite{cai2022memot}. Blatter et al. proposed Exemplar Attention to reduce runtime by sharing exemplar values among objects \cite{blatter2021efficient}. Zeng et al. used Query Interaction Modules and tracklet-aware label assignment to filter irrelevant queries and encourage one-to-one object-query mapping \cite{zeng2021motr}.

Attention and memory substantially improve re-identification and reduce ID switches after occlusion. Memory design, selection policies, and distractor handling are practical challenges, and attention modules can increase latency. Attention mechanisms are frequently embedded in transformer pipelines but also provide modular benefits in detection+association and Siamese systems.

\subsection{Foundation Models}
Foundation models supply rich, transferable representations and enable open-vocabulary detection. They assume large-scale pretraining yields features that generalize to downstream MOT tasks with little adaptation.

These models replace or augment backbones and embedding heads with foundation model features such as DINOv2/3 or Grounding DINO. They use language grounding to detect arbitrary classes and combine dense features with trajectory management or SAM-based mask propagation to build segmentation-aware trackers. Check \autoref{tab:foundation} to find a brief summary of the works on MOT using foundation models.

\begin{table*}
\centering
\caption{Summary of foundation models in MOT}\label{tab:foundation}
\footnotesize
\begin{tabular}
    {C{0.03\textwidth}
  C{0.04\textwidth}
  L{0.41\textwidth}
  L{0.33\textwidth}}
\toprule
\textbf{Ref.} & \textbf{Year} & \textbf{Architecture} & \textbf{Dataset(s)}\\
\midrule

\cite{oquab2023dinov2} & 2023 & Self-supervised ViT backbone (DINO+iBOT-style objectives) & LVD-142M (pretraining set) \\
\cite{liu2024grounding} & 2024 & Language-grounded transformer detector (open-vocabulary) & COCO (zero-shot), LVIS, ODinW, RefCOCO \\
\cite{ravi2024sam} & 2024 & Promptable image+video segmentor (transformer with streaming memory) & SA-V (Segment Anything Video) \\
\cite{yang2024samurai} & 2024 & SAM2-based zero-shot tracker with motion-aware memory selection & LaSOT$_\text{ext}$, GOT-10k \\
\cite{simeoni2025dinov3} & 2025 & Self-supervised ViT with Gram anchoring (stable dense features) & ADE20k, VOC12, Cityscapes, NYUv2, KITTI \\
\cite{videnovic2025distractor} & 2025 & SAM2.1 tracker with distractor-aware memory + memory management (DAM4SAM) & VOT2020, VOT2022, DiDi, LaSOT , LaSOText, GOT-10k, VOTS2024 \\
\cite{jiang2025sam2mot} & 2025 & MOT-by-segmentation on SAM2 + trajectory manager + cross-object interaction & DanceTrack, UAVDT-MOT, BDD100K-MOT \\
\bottomrule
\multicolumn{4}{l}{{\textit{Note:} Most foundation-model papers do not report MOT HOTA. HOTA is shown only when explicitly reported for MOT.}}
\end{tabular}
\end{table*}

Dense pre-trained backbones provide patch-level features for both semantic and precise spatial signals. DINOv2 supplies dense features that can be used frozen or with light adaptation \cite{oquab2023dinov2}. DINOv3 extends dense representation stability with Gram Anchoring \cite{simeoni2025dinov3}. Grounding DINO brings language-conditioned zero-shot detection to MOT, enabling detection by natural language for categories not present in training data \cite{liu2024grounding}. SAM-based systems such as SAMURAI and SAM2MOT combine mask propagation with trajectory management to transform segmentation primitives into full MOT systems \cite{yang2024samurai, jiang2025sam2mot}. DAM4SAM adds distractor-aware memory to retain discriminative anchor frames for identity distinction \cite{videnovic2025distractor}.

Practical integration challenges include adapting heavy foundation backbones for real-time use via distillation, adapters, or partial freezing. Foundation models are powerful for generalization and open-vocabulary tasks but need careful adaptation to meet latency and domain-specific requirements.

Foundation models improve generalization and semantic tracking capabilities and reduce annotation needs. Their drawbacks are inference cost and domain adaptation requirements. They are complementary to motion, graph, and memory techniques and can be distilled for deployment.

\subsection{Siamese Network}
Siamese networks learn an embedding where same-object pairs are close and different-object pairs are distant. They assume appearance similarity, possibly combined with motion, suffices for association.

In Siamese Network-based MOT approaches, a twin encoder processes two inputs (patch-to-patch or detection-to-tracklet) with shared weights. A similarity score guides matching or post-hoc tracklet reconnection. Siamese architectures are often augmented with multi-scale features, attention, or motion modules. See \autoref{tabsiam} for a summary of recent works on MOT that utilize Siamese Network.

\begin{table*}
\centering
    \caption{Summary of Siamese Network based approaches}
    \label{tabsiam}
\footnotesize
    \begin{tabular}
    {C{0.03\textwidth}
  C{0.04\textwidth}
  L{0.41\textwidth}
  L{0.23\textwidth}
  L{0.12\textwidth}}
    \toprule
    \textbf{Ref.} & \textbf{Year} & \textbf{Method} & \textbf{Dataset(s)} & \textbf{MOTA (\%)} \\
    \midrule
    \cite{xing2022siamese} & 2020 & CNN for Appearance extraction, LSTM and RNN for Motion modelling & Duke-MTMCT, MOT16 & 73.5, 55.0 \\
\cite{shuai2021siammot} & 2021 & Implicit and Explicit motion modelling & MOT17,  TAO-person, HiEve & 65.9, 44.3 (TAP@0.5), 53.2 \\    
\cite{gao2022multi} & 2021 & Siamese Network with Region Proposal Network & MOT16, MOT17, MOT20 & 65.8, 67.2, 62.3 \\
\cite{blatter2021efficient} & 2021 & Single instance level attention  & TrackingNet & 70.55 (Precision) \\
\cite{wan2022dsrrtracker} & 2022 & Dynamic search region refine and attention based tracking & MOT17, MOT20 & 67.2, 70.4 \\
\cite{xing2022siamese} & 2022 & Transformer based appearance similarity & UAV123  & 85.83 (Precision) \\
    \bottomrule
    \end{tabular}
\end{table*}




Xing et al. designed a Siamese Transformer Pyramid Network that integrates lightweight transformer attention into a multi-scale pyramid to improve robustness to scale variation and partial occlusion \cite{xing2022siamese}. Shuai et al. embedded a Siamese module into Faster R-CNN to produce efficient region-based matching \cite{shuai2021siammot}. Gao et al. proposed a Siamese Region Proposal Network as a prediction module with adaptive thresholding for stable matching \cite{gao2022multi}. Ma et al. used Siamese bidirectional GRUs to cleave corrupted tracklets and then reconnect fragments via a re-connection network \cite{ma2021deep}. Wan et al. and Blatter et al. introduced lightweight attention-augmented Siamese heads and exemplar transformers for efficient matching on resource constrained hardware \cite{wan2022dsrrtracker, blatter2021efficient}.



Siamese approaches are efficient and explicitly optimize matching. They struggle when appearance changes rapidly or discriminative features are absent. They pair well with motion priors, attention modules, or graph reasoning when appearance alone is insufficient.

\subsection{Tracklet Association}
Tracklet methods assume short-term local association is reliable and that higher-level linking of these short fragments yields robust long-term trajectories. This reduces sensitivity to noisy per-frame matches.

A typical workflow in this approach is that local association generates short tracklets. Tracklets are summarized by appearance and motion descriptors and linked at a higher level using temporal alignment, geometric constraints, or learned embeddings. Splitters, connectors, and memory banks are common submodules.

\begin{table*}
\centering
    \caption{Summary of Tracklet Association based approaches}
    \label{tabtracklet}
\footnotesize
    \begin{tabular}
    {C{0.03\textwidth}
  C{0.04\textwidth}
  L{0.41\textwidth}
  L{0.23\textwidth}
  L{0.12\textwidth}}
    \toprule
    \textbf{Ref.} & \textbf{Year} & \textbf{Method} & \textbf{Dataset(s)} & \textbf{MOTA (\%)} \\
    \midrule
    \cite{peng2020tpm} & 2020 &  Tracklet-plane matching process to resolve confusing short tracklets & MOT16, MOT17 & 50.9, 52.4 \\
\cite{ma2021deep} & 2020 & CNN for Appearance extraction, LSTM and RNN for Motion modelling & Duke-MTMCT, MOT16 & 73.5, 55.0 \\
\cite{stadler2021improving} & 2021 & Regression based two stage tracking &  MOT16, MOT17, MOT20 &  66.8,65.1,61.2 \\
\cite{wang2022split} & 2021 & Tracklet splitter splits potential false IDs and connector connects pure tracks to trajectory & MOT17, MOT20 & 61.5, 54.6 \\
\cite{nguyen2022lmgp} & 2021 & CenterTrack \cite{zhou2020tracking} and DG-Net \cite{zheng2019joint} as tracking graph and  GAEC+KLj \cite{keuper2015efficient} heuristic solver for lifted multicut solver & WILDTRACK, PETS-09, Campus & 97.1, 74.2, 77.5 \\
\cite{yu2022towards} & 2022 & Learnable view sampling for similarity-guided feature fusion and Trajectory-center memory bank for re-identification & MOT15, MOT16, MOT17, MOT20 & 62.1, 74.3, 73.5, 63.2 \\
    \bottomrule
    \end{tabular}

\end{table*}

Peng et al. introduced Tracklet-Plane Matching where short tracklets are aligned within temporal hyperplanes to enable matching across non-neighboring intervals \cite{peng2020tpm}. Ma et al. used Position Projection Networks to convert locally estimated trajectories into a global frame for better alignment \cite{ma2021deep}. Nguyen et al. proposed a 3D geometric formulation for multi-camera tracklet generation followed by joint spatio-temporal optimization \cite{nguyen2021lmgp}. Wang et al. developed TBooster, which splits tracklets at likely identity-change points and reconnects correct fragments using discriminative tracklet embeddings \cite{wang2022split}. Yu et al. presented MTCL, using trajectory memory banks and multi-view contrastive learning to compute robust association costs and mitigate ambiguity across views \cite{yu2022towards}. Stadler et al. apply regression-based re-identification with temporal direction cues to reidentify occluded objects \cite{stadler2021improving}.

Tracklet linking recovers long-term identities and is effective for long occlusions. Its success depends on the purity of initial tracklets; erroneous short tracklets propagate errors. Tracklet strategies complement graph-based global reasoning and memory mechanisms.

\subsection{Generative Tracking}
Generative approaches cast tracking as a stochastic refinement problem that models uncertainty and multi-modal hypotheses. They assume that iteratively denoising noisy proposals can resolve ambiguities better than deterministic regressors in crowded scenes.

These approaches sample noisy point-sets or bounding box hypotheses and iteratively denoise them via diffusion models that jointly refine locations and identity assignments; optionally augment denoising with discriminative matching to stabilize identities \cite{xie2024diffusiontrack, hu2025diffusionmot}. A summary of recent works on MOT that utilize Generative Tracking are shown in \autoref{tab:generative}.

\begin{table*}
\centering
\caption{Summary of generative (diffusion-based) tracking approaches}\label{tab:generative}
\footnotesize
\begin{tabular}
{C{0.03\textwidth}
  C{0.04\textwidth}
  L{0.41\textwidth}
  L{0.23\textwidth}
  L{0.12\textwidth}}
\toprule
\textbf{Ref.} & \textbf{Year} & \textbf{Architecture} & \textbf{Dataset(s)} & \textbf{HOTA (\%)} \\
\midrule
\cite{xie2024diffusiontrack} & 2024 &
DDPM-style proposal refinement; models box(-pair) relationship / point-set box representations &
MOT17, MOT20, DanceTrack &
60.8, 55.3, 52.4 \\
\cite{lv2024diffmot} & 2024 &
Decoupled diffusion-based motion predictor (D2MP) for non-linear motion &
DanceTrack, SportsMOT, MOT17, MOT20 & 62.3, 76.2, 64.5, 61.7 \\
\cite{liu2024pro2diff} & 2024 &
Deformable-DETR baseline; diffusion used during training & MOT17, DanceTrack & 57.6, 61.9 \\
\cite{hu2025diffusionmot} & 2025 & DiffusionDet-style detector + Pair-based Two-stage Match & DanceTrack, SportsMOT, MOT20, MOT17 & 65.3, 63.3 \\
\bottomrule
\end{tabular}

\end{table*}

DiffusionTrack formulates MOT as denoising point-sets, removing reliance on heavy heuristics by jointly refining detection and association \cite{xie2024diffusiontrack}. DiffMOT and Pro2Diff specialize diffusion for motion prediction and proposal propagation, reducing sampling steps or decoupling conflicting objectives \cite{lv2024diffmot, liu2024pro2diff}. DiffusionMOT augments diffusion with IoU+ReID matching and parallel sampling to reduce ID switches and speed inference \cite{hu2025diffusionmot}. Diffusion-based motion predictors learn distributions over future states to better capture multi-modal motion patterns in crowded or sports scenes \cite{lv2024diffmot}.

Generative trackers naturally express uncertainty and handle multi-modal outcomes. They perform well in ambiguous scenes but are computationally demanding and require design choices to ensure identity stability across denoising steps. They can integrate motion priors and memory to improve stability.

\begin{table*}
\centering
\caption{Summary of Mamba-based approaches}\label{tab:mamba}
\footnotesize
    \begin{tabular}
    {C{0.03\textwidth}
  C{0.04\textwidth}
  L{0.41\textwidth}
  L{0.23\textwidth}
  L{0.12\textwidth}}
    \toprule
	\textbf{Ref.} & \textbf{Year} & \textbf{Architecture} & \textbf{Dataset(s)} & \textbf{HOTA (\%)} \\
	\midrule
	   \cite{xiao2024mambatrack} & 2024 & Mamba Motion Predictor (MTP) with Tracklet Patching & DanceTrack, SportsMOT & 56.8, 72.6 \\
\cite{huang2025mambamot} & 2025 & Mamba-based Motion Predictor (Kalman Filter Replacement) & DanceTrack, SportsMOT & 56.1, 71.3 \\
\cite{liu2025mambavlt} & 2025 & Time-Evolving Multimodal State Space Framework (HMSS + SLE) & TNL2K, LaSOT, OTB99 & - \\
	\bottomrule
	\end{tabular}

\end{table*}

\subsection{Mamba-Based Tracking}
State-space models such as Mamba provide linear-complexity sequence modeling that enables long-horizon temporal reasoning with low latency. They assume the sequence structure can be captured by parameterized state-space kernels.

This method replaces or augments attention-heavy temporal modules with SSM/Mamba layers for bidirectional temporal encoding and motion prediction. It integrates SSM outputs with appearance embeddings and association heads to exploit long context efficiently. A summary of the mamba-based MOT approaches are shown in \autoref{tab:mamba}.

MambaTrack and MambaMOT use bidirectional Mamba predictors to interpolate missing observations under occlusion and to handle non-linear motion while keeping inference cost low \cite{xiao2024mambatrack, huang2025mambamot}. MambaVLT models visual and linguistic states jointly with selective scanning, enabling multimodal long-context modeling for tasks such as referring-expression conditioned tracking \cite{liu2025mambavlt}. SSMs provide long-range context at lower cost than full attention but require careful kernel parameterization to capture complex scene dynamics.

SSMs offer long temporal context with linear complexity and integrate with motion, graph, or transformer backbones. Their effectiveness depends on kernel design and their capacity to represent highly non-linear dynamics. In extreme interaction-rich scenes, full attention-based models may still have advantages.

Across these families we observe clear complementarities. Detection and association pipelines remain the engineering backbone for deployment. Motion and SSM models supply essential physical priors. Transformers and graph-based models provide global reasoning and learnable association rules. Attention and memory strengthen re-identification across occlusions. Foundation models boost generalization and enable semantic tracking. Siamese metrics yield efficient matching. Tracklet-level linking recovers long trajectories and diffusion models provide principled uncertainty handling. When choosing a strategy we recommend the following pragmatic rule of thumb. Prioritize robust detection and candidate preservation for real-world scenes. Add motion or geometry priors when camera or object dynamics are structured. Introduce memory or graph reasoning when long-term occlusion is frequent. Reserve heavy transformer or foundation-model deployments for scenarios where compute and data permit substantial gains in global reasoning or open-vocabulary capability.

\begin{figure*}
    \centering
    \includegraphics[width=\textwidth]{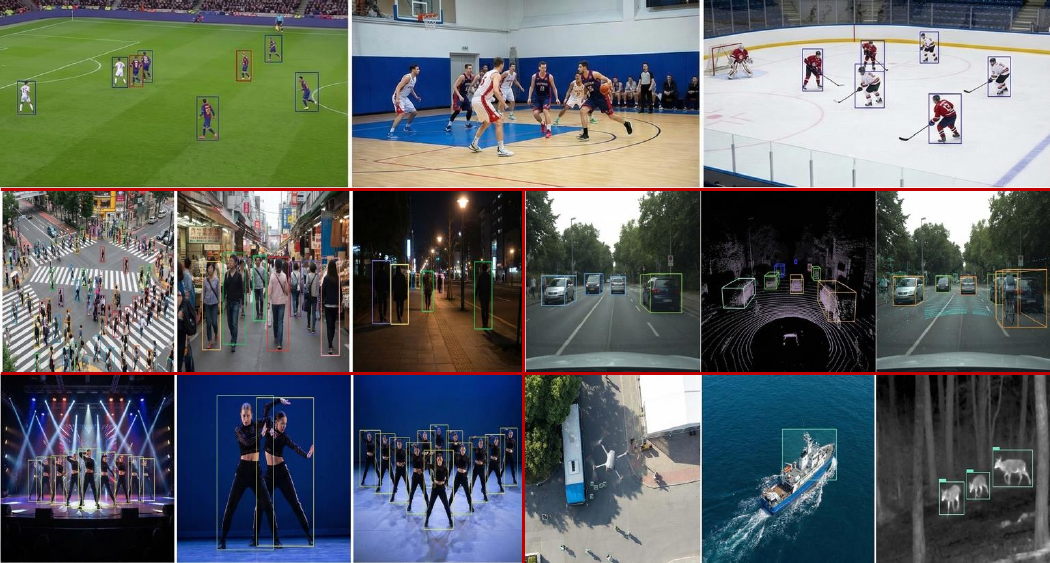}
    \caption{Visualization of diverse Multiple Object Tracking (MOT) benchmark datasets spanning various domains. The examples include sports analytics (top row), urban traffic and autonomous driving across different modalities and lighting (middle row), and complex scenarios such as dance tracking, dense crowds, maritime surveillance, and wildlife monitoring (bottom row).}
    \label{fig:datasetvisualization}
\end{figure*}

\section{Benchmark Datasets: The New Standards}\label{sec:datasets}

Benchmarks and standardized datasets are the backbone of empirical progress in MOT. They provide common inputs, shared annotation conventions, and reproducible metrics so that researchers can fairly compare methods, isolate failure modes, and measure real improvements over time. Well-designed datasets also expose gaps in existing algorithms by stressing specific factors (crowding, appearance ambiguity, fast motion, multi-modality), thereby guiding methodological innovation. We refer the reader to \autoref{tab:datasets} for a compact comparison.

\begin{table*}
\caption{Statistics of publicly available MOT datasets}
\label{tab:datasets}
\centering
\footnotesize
\begin{tabular}{c c l c c}
\hline
\textbf{Ref.} & \textbf{Year} & \textbf{Name} & \textbf{Frame Count} & \textbf{Size (Bytes)}\\
\hline
\cite{andriluka2008people} & 2008 & TUD Campus, Crossing       & 272           & 100M      \\
\cite{ferryman2009pets2009} & 2009 & PETS 2009                  & --            & 4.9G      \\
\cite{andriluka2010monocular} & 2010 & TUD Multiview Pedestrians  & 179           & 387M      \\
\cite{geiger2012we} & 2012 & KITTI Tracking             & --            & 15G       \\
\cite{leal2015motchallenge} & 2015 & MOT15                      & 11{,}283      & 1.3G      \\
\cite{milan2016mot16} & 2016 & MOT16                      & 11{,}235      & 1.9G      \\
\cite{milan2016mot16} & 2016 & MOT17                      & 33{,}705      & 5.5G      \\
\cite{patino2016pets} & 2016 & PETS 2016                  & --            & --        \\
\cite{voigtlaender2019mots} & 2019 & MOTS                       & 5{,}906       & 783.5M    \\
\cite{dendorfer2020mot20} & 2020 & MOT20                      & 13{,}410      & 5.0G      \\
\cite{pedersen20203d} & 2020 & 3D-ZeF20                   & 14{,}398      & 14.0G     \\
\cite{dave2020tao} & 2020 & TAO                        & 880{,}000     & 347G      \\
\cite{anjum2020ctmc} & 2020 & CTMC-v1                    & 152{,}498     & 768M      \\
\cite{liang2024ovt} & 2020 & OVT-B                      & 88{,}000      & 350G      \\
\cite{hu2022processing} & 2020 & Waymo Open Dataset         & --            & 336.6G    \\
\cite{voigtlaender2021reducing} & 2021 & TAO-VOS                    & --            & 2.4G      \\
\cite{sundararaman2021tracking} & 2021 & Head Tracking 21           & 11{,}464      & 4.1G      \\
\cite{weber2021step} & 2021 & STEP-ICCV21                & 2{,}075       & 380M      \\
\cite{fabbri2021motsynth} & 2021 & MOTSynth-MOT (CVPR22)      & 1{,}381{,}119 & --        \\
\cite{fabbri2021motsynth} & 2021 & MOTSynth-MOTS (CVPR22)     & 1{,}378{,}244 & --        \\
\cite{doering2022posetrack21} & 2022 & PoseTrack21                & 46{,}933      & --        \\
\cite{Sun_2022_CVPR} & 2022 & DanceTrack$^{*}$           & 105{,}000     & 16.5G     \\
\cite{caesar2020nuscenes} & 2020 & nuScenes                   & 1{,}400{,}000 & --        \\
\cite{cui2023sportsmot} & 2023 & SportsMOT$^{\dagger}$      & 150{,}000     & --        \\
\cite{scott2024teamtrack} & 2024 & TeamTrack$^{\ddagger}$     & 279{,}900     & --        \\
\cite{li2025mmot} & 2025 & MMOT                       & 13{,}000      & --        \\
\cite{barnard2025deepsea} & 2025 & DeepSea MOT                & 2{,}400       & --        \\
\hline
\multicolumn{5}{l}{$^{*}$Indoor scenes only.} \\
\multicolumn{5}{l}{$^{\dagger}$Includes basketball, volleyball, and football.} \\
\multicolumn{5}{l}{$^{\ddagger}$Includes soccer, basketball, and handball with 4.3M bounding boxes.}
\end{tabular}
\end{table*}
\subsection{General and Pedestrian Tracking Benchmarks}

MOT17 and MOT20 remain the community’s canonical pedestrian benchmarks. MOT17 contains multiple sequences captured with static and moving cameras and annotated with 2D bounding boxes and identity labels; MOT20 focuses on extremely crowded pedestrian scenes and provides dense annotations across 8 sequences (four train / four test) totaling 13,410 frames. Both datasets standardize per-frame bounding-box ground truth and identity trajectories, which has helped stabilize evaluation protocols and leaderboard comparisons \cite{dendorfer2020mot20}.

These benchmarks have driven incremental engineering gains and diagnostic practices (e.g., per-sequence analysis of ID switches and fragmentation). However, they show signs of metric saturation: many modern systems report MOTA and IDF1 in high ranges (reported saturation trends are discussed in \cite{hashempoor2025fasttracker}), which reduces their power to discriminate new ideas that target harder failure modes. MOT20’s crowded sequences still serve as an important stress-test for association under overlap, but both datasets are limited in domain diversity (pedestrian-centric, relatively short sequences) and in exposing challenges such as near-identical appearances or extreme long-term occlusions.

\subsection{Appearance-challenging Data: DanceTrack}
DanceTrack was designed to reveal an appearance-bias in classical MOT datasets by using dancers who frequently wear similar clothing and execute complex, non-linear motions \cite{Sun_2022_CVPR}. Its scale (105k frames over 100 videos) and content intentionally make per-frame detection comparatively easy while rendering association hard: appearance cues are weak and motion is complex. As a result, DanceTrack has shifted attention toward motion-centric and long-horizon association models (for example, MambaTrack and other state-space or learned-motion trackers that prioritize temporal reasoning) and highlighted that detection improvements alone do not solve association problems \cite{xiao2024mambatrack}. The dataset therefore serves as a targeted benchmark to evaluate temporal modeling and re-identification robustness under appearance ambiguity. Its limitation is clear: DanceTrack is domain-specific (performances) and does not cover other real-world factors such as varied lighting, sensor modalities, or multi-view setups.

\subsection{Sports Analytics Benchmarks}

Sports datasets expose high-velocity motion, strong inter-instance appearance similarity (team uniforms), and frequent occlusions, properties that are rare in pedestrian street scenes. SportsMOT provides 240 sequences, $>$150k frames, and 1.6M bounding boxes across basketball, volleyball and football, with dedicated splits for train/val/test to support robust benchmarking \cite{cui2023sportsmot}. Its principal stressors are fast, abrupt, and non-linear motion plus uniform appearances that break simple re-ID assumptions; these characteristics have driven the use of learned motion models (LSTM, transformer predictors) and domain-specific heuristics in recent trackers.

TeamTrack complements SportsMOT with overhead/bird’s-eye perspectives for tactical analysis, introducing other challenges such as strong scale variation and unique occlusion geometry \cite{scott2024teamtrack}. Together, sports benchmarks force algorithms to marry fine-grained motion forecasting with robust temporal association and to consider application-specific constraints (tracking frequency, downstream analytics).

\subsection{Autonomous Driving and 3D MOT Benchmarks}

Autonomous driving produces arguably the most demanding MOT requirements because trackers must operate across sensors, at long ranges, under safety-critical constraints. KITTI provided an early standardized benchmark for vehicle and pedestrian tracking \cite{geiger2013vision}. nuScenes scaled the problem to 360° sensing and multi-modality (camera, LiDAR, radar) with far larger data volume (1.4M frames) and richer annotations \cite{caesar2020nuscenes}. The Waymo Open Dataset extended scale and diversity further with multi-view, geographic and weather variation \cite{hu2022processing}. These datasets stimulated 3D MOT, sensor fusion, and motion-centric evaluation (e.g., evaluating velocity/acceleration accuracy in addition to spatial association).


MCTrack is an effort to unify evaluation across KITTI, nuScenes and Waymo by normalizing coordinate systems and detection formats (the BaseVersion), enabling cross-dataset evaluation without onerous preprocessing \cite{wang2024mctrack}. MCTrack also introduced motions-focused metrics (velocity and acceleration accuracy) that align evaluation more closely with downstream tasks in planning and prediction. While these testbeds are large and realistic, challenges remain: annotation heterogeneity across datasets, the computational cost of multi-modal methods, and the need for standardized open-source evaluation tools—gaps that MCTrack explicitly attempts to reduce.
\subsection{Emerging and Specialized Benchmarks}
Beyond the major categories, the community has produced many targeted datasets: UAV/aerial tracking, head-cropped datasets for fine-grained pedestrian analysis, dense crowd datasets, multi-camera network datasets, and low-light / adverse-condition collections (some listed in \autoref{tab:datasets}). These benchmarks are invaluable because they isolate single factors (viewpoint, altitude, lighting, cross-camera continuity) and help expose weaknesses of otherwise high-performing trackers. The proliferation of specialized datasets encourages cross-benchmark evaluation practices: modern works increasingly report results across multiple, diverse benchmarks to demonstrate robustness and generalization.


\section{MOT Metrics}\label{sec:metrics}
Evaluation in multiple object tracking (MOT) is inherently multi-faceted: a tracker must detect which objects are present, localize them precisely in each frame, and maintain consistent identities over time despite occlusion, missed detections and varying track lengths. Because different metrics emphasize different failure modes, fair and informative evaluation requires a small set of complementary measures that separate detection, localization and association performance. A summary of the evaluation metrics used in MOTA is given in \autoref{tab:eval}.

\begin{table*}[htb]
    \centering
    \caption{A summary of Multi-Object Tracking Metrics}
    \label{tab:eval}
    \footnotesize
    \begin{tabular}{L{.1\textwidth}L{.25\textwidth}L{.25\textwidth}L{.25\textwidth}}
        \toprule
        \textbf{Metric} & \textbf{Description} & \textbf{Pros} & \textbf{Cons}\\
        \midrule
        MOTP & Measures average localization error (e.g., IoU) for matched detections & Quantifies spatial precision of object localization & Ignores detection (misses/false positives) and association consistency\\
        MOTA & Aggregates misses, false positives, and ID switches into a single score & Widely used standard; correlates well with detection quality & Biased toward detection; underpenalizes ID switches; can be negative; penalizes error recovery\\
        IDF1 & F1 score based on the longest unique bijective mapping of trajectories & Emphasizes long-term identification and association consistency & Can penalize better detection if association is suboptimal; incentivizes long tracks over precision\\
        Track-mAP & Extends mAP to trajectories based on similarity thresholds & Unifies detection and tracking confidence; accounts for precision/recall & Difficult to visualize/interpret; susceptible to manipulation via low-confidence tracks\\
        HOTA & Unified metric balancing detection, association, and localization & Decomposable into sub-metrics (DetA, AssA, LocA); balances errors equally & More complex calculation; newer standard with less historical comparison data\\
        MOTSA & Extension of MOTA for pixel-level segmentation masks & Evaluates tracking and segmentation simultaneously & Inherits MOTA's detection bias and sensitivity to thresholds\\
        AMOTA & Averages MOTA scores across different recall thresholds & Robust to confidence threshold selection; useful for diverse datasets & Can obscure performance at specific operating points\\
        Cell-HOTA & Extension of HOTA for cell tracking, adding division accuracy & Explicitly evaluates lineage consistency and mitosis events & Domain-specific; requires specialized lineage ground truth\\
        Motion-Centric & Metrics measuring velocity angle, norm, and inversion errors & Critical for autonomous driving safety; assesses motion dynamics & Requires velocity ground truth; specific to motion prediction applications\\
        \bottomrule
    \end{tabular}
\end{table*}

\subsection{Traditional Metrics}

\subsubsection{Multiple Object Tracking Precision (MOTP)}
MOTP quantifies the localization precision of a tracker, specifically measuring the alignment between matched ground truth and hypothesis positions \cite{bernardin2008evaluating}. It averages the overlap error $d^i_t$ over all matches $c_t$:
$$MOTP = \frac{\sum_{i,t} d^i_t}{\sum_t c_t}$$

While MOTP assesses how well a tracker locates objects, it ignores detection performance (misses/false positives) and association consistency. Consequently, it is almost always reported alongside MOTA to provide a complete performance picture.

\subsubsection{Multiple Object Tracking Accuracy (MOTA)}

MOTA remains one of the most widely used metrics. It aggregates three error types: misses ($m_t$), false positives ($fp_t$), and identity switches ($mme_t$), normalized by the number of ground truth objects $g_t$ \cite{bernardin2008evaluating}:
$$MOTA = 1 - \frac{\sum_{t} (m_t + fp_t + mme_t)}{\sum_t g_t}$$

Despite its popularity, MOTA has significant limitations. It heavily biases evaluation toward detection performance while under-penalizing association errors (ID switches). Furthermore, it ignores association quality beyond first-order matching, allows for unbounded negative scores, and penalizes trackers that correct associations (counting the correction as a switch).



\subsubsection{The Identification Metric (IDF1)}


To address MOTA's insensitivity to temporal consistency, the IDF1 calculates an F1-score based on trajectory-level bijective mapping rather than frame-by-frame matching \cite{luiten2021hota}. It utilizes Identity Recall and Precision:
\begin{align*}
\text{ID-Recall} &= \frac{\lvert \text{IDTP} \rvert}{\lvert \text{IDTP} \rvert + \lvert \text{IDFN} \rvert}\\
\text{ID-Precision} &= \frac{\lvert \text{IDTP} \rvert}{\lvert \text{IDTP} \rvert + \lvert \text{IDFP} \rvert}\\
\text{IDF1} &= \frac{\lvert \text{IDTP} \rvert}{\lvert \text{IDTP} \rvert + 0.5\lvert \text{IDFP} \rvert + 0.5\lvert \text{IDFN} \rvert}
\end{align*}

Here, $IDTP$, $IDFN$, and $IDFP$ represent identity true positives, false negatives, and false positives, respectively.

While IDF1 effectively emphasizes long-term association, it can produce lower scores for trackers with high detection accuracy if the trajectory mapping is suboptimal, creating a trade-off where trackers are incentivized to maintain long tracks even at the cost of spatial precision.

\subsubsection{Track-mAP}
Track-mAP extends the mean Average Precision metric to trajectories. A predicted trajectory matches a ground truth if their similarity $S_{tr}$ exceeds a threshold $\alpha_{tr}$ \cite{luiten2021hota}. Precision ($Pr_n$) and Recall ($Re_n$) are calculated at rank $n$:
\begin{align*}
Pr_n &= \frac{\lvert TPTr \rvert_n}{n}\\
Re_n &= \frac{\lvert TPTr \rvert_n}{\lvert gtTraj \rvert}\\
\end{align*}
The final score is the area under the interpolated precision-recall curve:
$$InterpPr_n = \max_{m \geq n} (Pr_m)$$
Although Track-mAP unifies detection and tracking confidence, it is difficult to interpret and visualize. Furthermore, it is susceptible to manipulation, where numerous low-confidence predictions can artificially inflate the score, obscuring the true quality of the tracking.

\subsection{Higher Order Tracking Accuracy (HOTA)}

HOTA addresses the imbalance between detection and association found in MOTA and IDF1. It geometrically averages detection and association scores, ensuring both are equally weighted \cite{luiten2021hota}. HOTA is calculated at various localization thresholds $\alpha$ and integrated:
$$HOTA_\alpha = \sqrt{\frac{\sum_{c \in {TP}} A(c)}{\lvert TP \rvert + \lvert FN \rvert + \lvert FP \rvert}}$$
with per-detection association accuracy
$$A(c) = \frac{\lvert TPA(c) \rvert}{\lvert TPA(c) \rvert + \lvert FNA(c) \rvert + \lvert FPA(c) \rvert}$$
The term $A(c)$ represents the association accuracy for a specific matched detection $c$:
$$A(c) = \frac{\lvert TPA(c) \rvert}{\lvert TPA(c) \rvert + \lvert FNA(c) \rvert + \lvert FPA(c) \rvert}$$
HOTA introduces novel concepts for error analysis: True Positive Associations ($TPA(c)$) are matches sharing correct IDs; False Negative Associations ($FNA(c)$) are missed associations (fragmentations); and False Positive Associations ($FPA(c)$) are incorrect associations (mergers/switches):
\begin{align*}
TPA(c) = \{k\} \mid k \in& \{TP \mid prID(k) = prID(c)\\ 
&\wedge gtID(k) = gtID(c)\}\\
FNA(c) = \{k\} \mid k \in& \{TP \mid prID(k) \neq prID(c)\\ 
&\wedge gtID(k) = gtID(c)\} \\
&\cup \{FN \mid gtID(k) = gtID(c)\}\\
FPA(c) = \{k\} \mid k \in& \{TP \mid prID(k) = prID(c)\\
&\wedge gtID(k) \neq gtID(c)\} \\
&\cup \{FP \mid prID(k) = prID(c)\}
\end{align*}

The final HOTA score approximates the integral over $\alpha$:
\begin{align*}
HOTA &= \int_{0}^{1} HOTA_\alpha , d\alpha \\
&\approx \frac{1}{19} \sum_{\alpha \in {0.05, 0.10, \ldots, 0.95}} HOTA_\alpha
\end{align*}

A key strength of HOTA is its decomposability into interpretable sub-metrics, allowing researchers to isolate specific error types:

\subsubsection{Localization Accuracy (LocA)}

LocA measures spatial alignment (average IoU) of matched detections \cite{luiten2021hota}.

\begin{align*}
    LocA = \int_{0}^{1} \frac{1}{\lvert TP_\alpha \rvert} \sum_{c \in \{TP_\alpha\}} S(c) \, d\alpha
\end{align*}
Unlike MOTP, this integrates over multiple thresholds.

\subsubsection{Detection Accuracy (DetA)}

DetA is the harmonic mean of Detection Recall ($DetRe$) and Detection Precision ($DetPr$), balancing misses and false positives.
$$DetA_\alpha = \frac{\lvert TP \rvert}{\lvert TP \rvert + \lvert FN \rvert + \lvert FP \rvert}$$










\subsubsection{Association Accuracy (AssA)}

AssA is the harmonic mean of Association Recall ($AssRe$) and Association Precision ($AssPr$). This quantifies how well the tracker maintains identities, penalizing fragmentations ($AssRe$ errors) and mergers ($AssPr$ errors).
$$AssA_\alpha = \frac{1}{\lvert TP \rvert} \sum_{c \in \{TP\}} A(c)$$









\subsection{Segmentation and Robustness Metrics}
\subsubsection{Multi-Object Tracking and Segmentation Accuracy (MOTSA)}

MOTSA extends MOTA to pixel-level masks \cite{voigtlaender2019mots}. It replaces bounding box inputs with mask overlap calculations:
$$MOTSA = 1 - \frac{\lvert FN \rvert + \lvert FP \rvert + \lvert IDS \rvert}{\lvert M \rvert}$$
where $M$ represents the set of ground truth masks. Like MOTA, it remains sensitive to matching thresholds and emphasizes detection over association.

\subsubsection{Average Multiple Object Tracking Accuracy (AMOTA)}

AMOTA improves upon MOTA's sensitivity to confidence thresholds by averaging MOTA across a range of recall values $r$:

\begin{align*}
    AMOTA = \frac{1}{L} \sum_{r \in \{\frac{1}{L}, \frac{2}{L}, \ldots, 1\}} \left(1 + \frac{FP_r + FN_r + IDS_r}{num\_gt}\right)
\end{align*}

where:
\begin{itemize}
    \item $num\_gt$: The total number of ground truth objects across all frames
    \item $FP_r$, $FN_r$, $IDS_r$: The number of false positives, false negatives, and identity switches for recall value $r$
    \item $L$: The total number of recall values evaluated
\end{itemize}

This provides a more robust, operating-point-agnostic evaluation, particularly useful for comparing trackers in diverse environments like autonomous driving.


\subsection{Domain-Specific and Emerging Metrics}
As MOT expands into specialized fields, generic metrics often fail to capture domain-critical requirements.

\subsubsection{Cell-HOTA (Biological Imaging)}
In cell tracking, capturing mitosis is essential. Cell-HOTA extends HOTA with \textit{Division Accuracy (DivA)}, which explicitly evaluates lineage consistency \cite{o2025cell}. While standard HOTA measures trajectory overlap, DivA validates that parent-daughter relationships are correctly identified, preserving the biological fidelity required for developmental research.

\subsubsection{Motion-Centric Metrics for Autonomous Driving}

Safety-critical applications require precise motion estimation beyond simple localization. The MCTrack framework \cite{wang2024mctrack} introduces three metrics to audit velocity dynamics:
\begin{enumerate}
\item Velocity Angle Error (VAE): Measures directional accuracy, critical for trajectory planning.
\begin{align*}
    VAE = \arccos\left(\frac{\vec{v}_{\text{pred}} \cdot \vec{v}_{\text{gt}}}{\|\vec{v}_{\text{pred}}\| \|\vec{v}_{\text{gt}}\|}\right)
\end{align*}
\item Velocity Norm Error (VNE): Captures speed estimation errors, which impact collision avoidance.
 \begin{align*}
    VNE = \|\vec{v}_{\text{pred}}\| - \|\vec{v}_{\text{gt}}\|
\end{align*}
\item Velocity Inversion Ratio (VIR): A binary metric flagging catastrophic failures where predicted motion opposes actual motion ($\vec{v}_{\text{pred}} \cdot \vec{v}_{\text{gt}} < 0$).
\begin{align*}
    VIR = \frac{\text{Count}(\vec{v}_{\text{pred}} \cdot \vec{v}_{\text{gt}} < 0)}{\text{Total velocity predictions}}
\end{align*}
\end{enumerate}

\subsubsection{Open-World and Variable Frame Rate Metrics}
To support deployment in uncontrolled environments, recent protocols have shifted toward class-agnostic evaluation. These metrics assess performance on novel objects without penalizing the tracker for categories unseen during training, and accommodate variable frame rate inputs, moving beyond the fixed-class, fixed-FPS assumptions of traditional benchmarks.

\section{Applications of Multiple Object Tracking}\label{sec:applications}

Multiple object tracking (MOT) now underpins a wide range of deployed and research systems. In this section we first outline the principal sectors that rely on MOT and then, for each sector, concisely state the task objectives, the dominant operational constraints (e.g., accuracy, latency, robustness, scalability), and deployment realities that shape tracker design and evaluation. \autoref{tab:applications} summarizes domains, typical targets, and the tracking priorities they impose,

\begin{table*}[htb]
    \centering
    
    \caption{Compact mapping of MOT application domains to typical targets, dominant sensors, the tracking priorities those domains impose, representative benchmarks/methods cited in the text, and common deployment constraints}
    \label{tab:applications}
    \footnotesize
    \begin{tabular}{L{1.8cm} L{2.5cm} L{2.5cm} L{5cm} L{2cm}}
        \toprule
        \textbf{Domain} & \textbf{Typical Targets} & \textbf{Dominant Sensors} & \textbf{Primary Tracking Priorities} & \textbf{Representative Methods}\\
        \midrule
        Autonomous driving & Vehicles, pedestrians, cyclists, road infrastructure & Camera, LiDAR, radar & Real-time operation; kinematic fidelity (velocity/accel); occlusion resilience; sensor fusion; safety & \cite{wang2024mctrack, gao2020multiple,fu2021real,pang20213d}\\
        Pedestrian \& Re-ID & Pedestrians (single- and cross-camera) & CCTV, thermal, multi-camera networks & Identity persistence; cross-camera association; occlusion resilience; low-light operation & \cite{zhang2021online,sundararaman2021tracking,stadler2021improving}\\
        Vehicle surveillance & Vehicles & Roadside cameras, aerial platforms & Long-term tracks and global coordinate consistency; motion modeling for anomaly detection; speed estimation & \cite{shi2021anomalous,quang2021vietnamese,wang2021track,jimenez2022multi}\\
        Sports tracking & Players, referees, ball & Broadcast and multi-view cameras & High motion fidelity; identity stability under similar appearance; robustness to dense interactions; pose/interaction cues & \cite{cioppa2022soccernet,kalafatic2022multiple,naik2022deepplayer,vats2021player}\\
        Wildlife monitoring & Wild animals, livestock & UAV RGB, fixed cameras, occasionally thermal & Robustness to species variation and non-rigid motion; sparse labels; long-term behavior analysis & \cite{marcos2021animal,zhang2022animaltrack,guo2022video,ju2021turkey}\\
        Aerial/UAV surveillance & People, vehicles, animals, objects of interest & Aerial RGB, thermal, multispectral & Small apparent size handling; ego-motion compensation; low latency; multispectral fusion & \cite{li2025mmot}\\
        Biomedical \& healthcare & Surgical instruments, cells, anatomical landmarks & Endoscopic cameras, microscopy & Traceability and reliability; lineage preservation; event detection; high interpretability & \cite{nwoye2025cholectrack20,o2025cell}\\
        Marine \& aquatic & Fish, deep-sea fauna, aquatic organisms & Underwater video, fixed underwater cameras & Low-light robustness; non-rigid body handling; sparse targets; behavior inference & \cite{barnard2025deepsea,li2022cmftnet,dvechtverenko2022tracking}\\
        Precision agriculture & Individual plants, pests, livestock & Field cameras, UAVs & Per-plant/animal tracking; scalability; occlusion resilience; seasonal appearance variation & \cite{ge2022tracking,tan2022towards}\\
        Visual surveillance & People, objects, crowds & CCTV, thermal networks, multi-camera infrastructures & Long-duration operation; cross camera re-ID; forensic quality; privacy; low-light operation & \cite{ahmed2020towards,urbann2021online,nagrath2022understanding}\\
        Robotics \& HRI & Humans, manipulable objects, other robots & On-robot RGB/depth, audio-visual sensors & Real-time closed-loop tracking; low latency; safety; multimodal integration & \cite{wilson2020avot,pereira2022sort}\\
        \bottomrule
    \end{tabular}
\end{table*}

\subsection{Autonomous Driving and Traffic Management}
In autonomous driving, MOT is a safety-critical input to planning and control: vehicles must maintain continuous, multi-class tracks of pedestrians, cyclists, other vehicles and static infrastructure so that trajectory planners and collision-avoidance modules can act reliably in real time. This requirement pushes systems beyond 2D bounding-box association toward fused, 3D tracking across camera, LiDAR and radar. Work such as MCTrack standardizes 3D formats across KITTI, nuScenes and Waymo and importantly shifts evaluation toward motion, introducing velocity-centric metrics (VAE, VNE, VIR) that better reflect downstream safety needs \cite{wang2024mctrack}. Research trends reflect these operational pressures: dual-attention and self-attention architectures improve robustness under occlusion \cite{gao2020multiple,fu2021real}, random-finite-set–based filters add principled 3D motion handling \cite{pang20213d}, and LiDAR-first methods (e.g., point-cloud trackers such as SimTrack) produce more robust 3D associations in heavily occluded scenes. Fixed infrastructure for traffic monitoring shares many goals but emphasizes continuous, low-latency operation on resource-constrained edge hardware; lightweight detectors and efficient trackers are commonly deployed in such settings \cite{zou2022real,cho2022autonomous}. Practical systems therefore balance kinematic fidelity, sensor fusion complexity and the compute/bandwidth limits of edge deployments.

\subsection{Pedestrian Tracking and Person Re-Identification}

Pedestrian tracking exemplifies identity-centric MOT: the primary objective is to preserve identities across occlusions, low resolution, and view changes for surveillance, safety and analytics. By enabling continuous identification and tracking over time, MOT goes beyond gait recognition approaches \cite{hasan2022heatgait, hasan2023gaitgcn} that merely identify individuals without maintaining temporal continuity. The domain’s practical constraints, frequent occlusion, many near-identical appearances, cross-camera calibration errors and night-time operation, drive methods toward robust re-identification, head-centric detection and cross-camera trajectory matching rather than solely higher detection AP. Techniques such as DROP re-identify occluded pedestrians with appearance features \cite{zhang2021online}, while HeadHunter focuses on head detection plus re-identification to sustain identity under heavy occlusion \cite{sundararaman2021tracking}, and regression-based occlusion handling offers another path to stability \cite{stadler2021improving}. Thermal pipelines and cross-camera post-processing extend tracking to low-light conditions and multi-site deployments \cite{chen2021vehicle,ma2021deep,wang2022multi}. In practice, systems emphasize long-term identity persistence and robust cross-view association, trading off some detection-centric metrics to maintain the identity-level signals required by downstream analytics.

\subsection{Vehicle Surveillance and Anomaly Detection}
In vehicle surveillance the operational task is long-term monitoring and the detection of anomalous behaviour: unusual trajectories, speeding or other deviations from learned motion patterns. These use cases typically span large scenes and require global coordinate consistency, so researchers emphasize motion modeling, tracklet-to-track association and global motion priors. Motion-based models such as those using Gaussian local velocity capture typical vehicle movements to flag anomalies \cite{shi2021anomalous}; speed-estimation pipelines project detections into 3D and compute velocities for enforcement or analysis \cite{quang2021vietnamese}. Graph-based association and reconstruct-to-embed strategies convert local tracklets into reliable long-term tracks for downstream analysis \cite{wang2021track,zhang2022research}. Reviews focused on traffic MOT synthesize these approaches and the datasets used to evaluate them. In this context, \cite{jimenez2022multi} offers a comprehensive and informative overview.




\subsection{Sports Player Tracking and Analytics}
Sports analytics demands high-fidelity motion and identity information to extract tactics, workload and performance metrics. The challenges are specific: fast, non-linear player motion, often-homogeneous team apparel that weakens appearance cues, dense interactions, and broadcast-camera artifacts such as panning and zooming. These force trackers to trade off between motion fidelity and identity stability. SportsMOT, SoccerNet and sport-specific baselines have catalyzed methods tailored to these constraints \cite{cioppa2022soccernet,kalafatic2022multiple,naik2022deepplayer,vats2021player}. Practically, teams and broadcasters use MOT outputs for tactical analysis and to drive broadcast visual effects; recent work adapts large segmentation models (e.g., SAM 2) to produce pixel-accurate player masks that enable broadcast enhancements and biomechanical measurements without marker-based capture.

\subsection{Wildlife Tracking and Animal Monitoring}
MOT provides a non-invasive alternative to sensor-tagging for studying animal behaviour and population dynamics, but field conditions impose distinctive constraints: diverse morphologies, variable motion patterns, sparse labeled data and difficult viewpoints (UAV, ground cameras, underwater). Applied UAV systems combine lightweight detectors with particle-filter trackers to follow animals at scale \cite{marcos2021animal}, while benchmark datasets like AnimalTrack provide species-specific baselines to spur method development \cite{zhang2022animaltrack}. For livestock and welfare monitoring, trackers such as FairMOT and JDE have been adapted to monitor group behaviour of pigs and poultry \cite{guo2022video,ju2021turkey}. For a more comprehensive analysis of MOT techniques specifically tailored to livestock management and welfare, readers are referred to \cite{alanezi2022livestock}. Underwater domains add further complications (poor lighting, non-rigid bodies and schooling behaviour) prompting specialized architectures (e.g., CMFTNet) and domain surveys to consolidate progress \cite{li2022cmftnet,dvechtverenko2022tracking}.




\subsection{Unmanned Aerial Vehicles and Aerial Surveillance}
Aerial tracking deploys MOT from moving platforms, introducing camera ego-motion, small apparent target size and strict on-board compute limitations. Benchmarks for moving-camera MOT (e.g., MMOT) and multispectral sensing studies reflect the operational reality that aerial systems often combine visible, thermal and SAR data to remain robust under camouflage or adverse weather \cite{li2025mmot}. In search-and-rescue missions, thermal trackers materially extend operability into night and challenging terrain; practical systems must carefully trade detection range, latency and on-board processing to meet mission constraints.



\subsection{Biomedical and Healthcare Applications}

In biomedical settings MOT supports tasks ranging from surgical automation to quantitative cell biology, where requirements for reliability, interpretability and event preservation are stringent. Surgical-tool tracking (e.g., systems developed around CholecTrack) must handle fluid occlusions, specular highlights and tools leaving the field of view while delivering traceable outputs for workflow analysis and skill assessment \cite{nwoye2025cholectrack20}. In microscopy, trackers must preserve cell lineage and correctly detect division events; benchmarks such as Cell-HOTA explicitly measure Division Accuracy (DivA) to ensure that lineage and mitosis handling are assessed and preserved for downstream biological interpretation \cite{o2025cell}. These domains prioritize end-to-end traceability and integration with clinical or experimental pipelines over raw detection metrics.



\subsection{Marine Ecology and Aquatic Environments}
Automated tracking in marine ecology enables population estimation and behavioural analysis in environments that are hazardous or expensive for human observers. Deep-sea benchmarks (e.g., DeepSea MOT) and associated research address the particular problems of limited lighting, sparse targets and irregular morphologies that typify deep-water footage \cite{barnard2025deepsea}. Shallow- and fresh-water applications encounter schooling, vegetation occlusion and rapid appearance variation; practical systems and datasets have adapted evaluation protocols to these ecological idiosyncrasies \cite{li2022cmftnet}.



\subsection{Precision Agriculture and Crop Management}
In agriculture we apply MOT to per-plant or per-animal monitoring for phenotyping, pest detection and harvest optimization. Agricultural scenes are cluttered and seasonal appearance variation is large, so deployed systems favor lightweight detectors, multi-scale fusion and robust association to scale across fields under compute constraints. Case studies demonstrate practical pipelines, e.g., tomato and cotton monitoring using YOLO-family detectors coupled with trackers like DeepSORT, where the focus is on operational robustness and scalability rather than pushing laboratory detection benchmarks \cite{ge2022tracking,tan2022towards}.



\subsection{Visual Surveillance and Security}
Visual surveillance remains a foundational MOT application, now extending beyond simple motion alarms to long-term monitoring, crowd analysis and cross-camera re-identification in privacy-sensitive settings. Deployed systems must run continuously, handle cross-camera identity association and operate in low-light conditions, often using thermal imagery; benchmarks and challenges such as TP-MOT guide progress on nighttime and thermal tracking \cite{ahmed2020towards,urbann2021online,nagrath2022understanding}. Practical designs balance efficiency, privacy considerations and the need for forensic-quality outputs.



\subsection{Robotics and Human-Robot Interaction}

Robots rely on MOT for navigation, collaboration and multi-robot coordination; here tracking operates in closed-loop control stacks and must meet stringent latency and safety bounds. Integrating multi-modal cues (e.g., audio-visual tracking) and constructing cost matrices tailored to robotics tasks improves persistence and occlusion recovery in interactive settings \cite{wilson2020avot,pereira2022sort}. The practical emphasis is on real-time, reliable perception that can drive control decisions under uncertainty.


\section{Future Research Directions}\label{sec:future}
Multiple Object Tracking remains active and evolving; many of the limitations that characterize current systems directly motivate the research directions we expect to dominate in the near future. Common shortcomings, fragile cross-view identity persistence, brittle open-world generalization, limited 3D reasoning, excessive computational cost, sparse uncertainty estimates, and evaluation gaps, expose concrete opportunities for progress.

\subsection{Multi-Camera and Cross-Domain Tracking}

Tracking across camera networks, especially when views are non-overlapping, remains a practical bottleneck because identity must be maintained through large viewpoint, illumination and temporal gaps. Progress in registering scenes into common 3D coordinate systems can yield continuous geographic tracking useful for large-scale surveillance, airport operations and multi-venue sports; when feasible, projection into shared world coordinates reduces identity ambiguity \cite{ma2021deep}. Datasets and systems such as MMPTrack and real-time online multi-camera pipelines have advanced the state of the art \cite{han2021mmptrack,zhang2019real}, but deployment continues to be constrained by computational overhead and long-trajectory consistency. A promising conceptual shift is to combine visual signals with higher-level semantic descriptions—for example, leveraging visual-language and foundation models for re-identification—so that cross-camera association can use semantic context in addition to, or instead of, brittle appearance descriptors. This direction aims to trade pure appearance matching for richer, language-grounded identity cues that generalize across domains.


\subsection{Class-Aware and Open-Vocabulary Tracking}
Many applications require tracking particular classes or adapting to previously unseen categories, which exposes the limits of class-agnostic trackers. Open-vocabulary techniques and open-world tracking address this by integrating visual-language priors: methods such as OVTR use CLIP-style models as detection backbones to handle unseen categories, and EffOWT extends these ideas with parameter-efficient fine-tuning to scale to thousands of categories without full retraining. Going forward, we anticipate a stronger emphasis on few-shot and zero-shot approaches that allow trackers to adopt new class concepts from minimal supervision, and on semantic integration—using language-derived context to resolve ambiguous visual evidence. Trajectory-aware classification, where motion patterns inform category hypotheses under occlusion or low-visibility, is another natural extension.



\subsection{3D and Volumetric Tracking}
Shifting from 2D image tracks to 3D and volumetric representations addresses occlusion and localization weaknesses intrinsic to planar approaches. Depth-enabled tracking (via LiDAR point clouds, RGB-D streams, multi-view stereo reconstructions or spatiotemporal voxel/event data) permits explicit spatial reasoning, better kinematic estimates, and more reliable occlusion resolution. The MCTrack Framework is a concrete instantiation of the community push toward standardized 3D tracking in driving datasets \cite{wang2024mctrack}; yet applying comparable 3D rigor outside automotive domains and delivering real-time volumetric tracking remain open research problems. Efficiency, representation choice for volumetric data, and the extension of 3D benchmarks to non-automotive tasks are the core challenges to address.


\subsection{Advanced Architectural Paradigms}
Architectural innovation continues to reshape MOT. Transformers now realize end-to-end tracking formulations while approaching real-time budgets: single-decoder designs that encode temporal information directly in track queries reduce computation by eliminating redundant modules. Motion-aware transformer variants explicitly inject motion modeling into attention-based architectures and have shown strong performance on motion-centric benchmarks \cite{yang2025motion}. Parallel to transformers, State Space Models (SSMs) and Mamba-style sequence formalisms furnish linear-complexity temporal reasoning that scales to long histories without attention’s quadratic cost; they recast motion prediction as data-driven sequence modeling that captures non-linear dynamics. New paradigms like MOTIP, which treat MOT as an in-context identity prediction task, further challenge the distinction between association and detection by decoding identity labels directly for current detections \cite{gao2025multiple}. The immediate conceptual agenda is to explore hybrids that combine transformer flexibility with SSM efficiency and to develop adaptive computation strategies so model complexity scales with scene difficulty.



\subsection{Lightweight and Real-Time Architectures}
Real deployments demand compact models that meet strict latency and power budgets. Architectural simplification, multitask training, and inference-time optimizations (for instance as in OneTrack-M) can yield substantial speedups while preserving accuracy. Future work will increasingly leverage neural architecture search, distillation, quantization and pruning, and hardware–algorithm co-design to produce trackers that are both accurate and deployable on edge devices, ensuring real-time performance (targeting or exceeding practical thresholds such as 30 fps) in surveillance, robotics and mobile platforms.


\subsection{Quantum Computing for Optimization}
The association step in MOT is a combinatorial optimization problem that can become computationally intensive at scale. Quantum approaches, e.g., Adiabatic Quantum Computing formulations that map assignments to Ising models, have been explored as a potential path to faster combinatorial solvers \cite{zaech2022adiabatic}. As quantum hardware and hybrid classical–quantum algorithms mature, research should focus on formulating MOT-specific optimization instances for quantum solvers, hybrid pipelines that leave feature extraction classical while outsourcing hard assignment subproblems to quantum optimizers, and empirical benchmarking of quantum methods on realistic MOT instances.


\subsection{Language Models and Semantic Reasoning}
Multimodal and large language models open the door to reasoning-based tracking that leverages semantics and context in place of brittle visual descriptors. Language-grounded trackers could exploit object descriptions, scene semantics and commonsense relations to disambiguate identities, inform trajectory priors, or enable few-shot tracking from textual prompts. Conversational, human-in-the-loop interfaces for tracking correction and annotation, where language drives iterative refinement, are a practical application of this trend. More broadly, transferring temporal reasoning capabilities from language models into visual sequence understanding offers a route to richer, context-aware association and forecasting.


\subsection{Domain-Specific Extensions and Customization}
Domain requirements remain diverse and often non-overlapping: biomedical applications need lineage-preserving, division-aware trackers; marine ecology must handle non-rigid morphologies and sparse, low-contrast sightings; surgical tracking must contend with fluids, smoke, and tool deformation; sports analytics demands pose- and interaction-aware tracking. We expect continued growth in domain-specialized datasets, architectures and metrics that encode domain priors, rather than forcing a single generic tracker to perform adequately everywhere.


\subsection{Uncertainty Quantification and Robustness}
Most current trackers output point estimates with little calibrated uncertainty. Adding principled uncertainty quantification, via Bayesian methods, ensembling, or modern deep uncertainty techniques, would enable adaptive association thresholds, occlusion-aware propagation, principled detection of out-of-distribution scenes, and probabilistic trajectory forecasts with interpretable confidence bounds. Such capabilities are particularly important in safety-critical domains and in systems that combine automated tracking with human oversight.


\subsection{Dataset Trends and Gaps}
Datasets have driven methodological focus: saturation on classical benchmarks (MOT17/MOT20) shifted attention to motion-centric and appearance-challenging collections (DanceTrack, SportsMOT), to multi-modal/3D driving benchmarks (nuScenes, Waymo), and to segmentation- or open-vocabulary-aware tasks. Nevertheless, significant gaps persist: long-term identity persistence (hours or cross-camera), unified low-light/adverse-weather benchmarks, and standardized multi-dataset protocols that reward cross-domain generalization rather than leaderboard tuning. Annotation heterogeneity (boxes vs. masks, 2D vs. 3D coordinates, single-frame vs. long trajectories) further complicates fair comparison. We therefore recommend evaluation practices that disentangle detection and association failures, include at least one appearance-challenging and one multi-modal/3D benchmark (e.g., DanceTrack or SportsMOT plus nuScenes/Waymo), and report per-sequence, per-metric breakdowns (ID switches, fragmentation, velocity/acceleration errors) to make strengths and failure modes explicit.

\subsection{Adversarial Robustness and Safety}
As MOT moves into safety-critical deployments, adversarial robustness becomes a core research priority. We foresee systematic analysis of attack surfaces, robust training regimens, certified-defenses where possible, and runtime detection/mitigation strategies that can be integrated into real-time pipelines. Robustness work will need to tie closely to uncertainty estimation and to domain-aware risk metrics that map tracking errors to application-level consequences.

Taken together, the trends above point toward a field that will increasingly balance specialization with generalization. Foundation models and transfer learning reduce the data needed for new tasks; hybrid pipelines combine classical motion priors with emerging solvers; and end-to-end learnable components replace brittle heuristics where appropriate. Crucially, efficiency, deployability and rigorous uncertainty quantification will guide evaluation criteria as strongly as raw accuracy. These shifts should also drive changes in benchmarks and metrics and in dataset design to better reflect real-world constraints. MOT is therefore poised to evolve from benchmark-driven lab progress to systems-engineering advances that directly address application needs across domains.
\section{Conclusion}\label{s8}
Multi-Object Tracking remains a core perception challenge as systems transition from curated benchmarks to real-world deployment, where persistent identity under occlusion, dense interactions, and domain shift expose key limitations. This review shows that progress has arisen from complementary approaches rather than a single paradigm, spanning detection–association pipelines, end-to-end global reasoning models, motion and state-space priors, memory- and graph-based identity linking, and generative formulations for uncertainty. Foundation models are reshaping detection and representation learning, enabling transfer and open-vocabulary tracking while introducing new efficiency and adaptation trade-offs. We also highlight the strong influence of datasets and evaluation protocols on research directions: saturation of canonical pedestrian benchmarks has driven the emergence of appearance-challenging and multimodal datasets, particularly in autonomous driving and sports, favoring motion-aware and domain-adaptive designs. Despite this progress, a gap persists between leaderboard performance and deployment needs, including long-term identity stability, robustness under adverse conditions, real-time operation, and evaluation aligned with safety and downstream tasks.

Looking forward, unified evaluation standards that span domains and incorporate motion- and safety-aware metrics are critical. Promising directions include domain-adaptive pipelines that combine foundation-model representations with efficient temporal reasoning, integration of semantic and language cues for task-aware tracking, and principled uncertainty modeling to reduce catastrophic failures. We advocate shifting emphasis from isolated benchmark gains toward robustness, diagnostics, and deployment-aware evaluation, providing a structured roadmap for the next phase of MOT research.

\section*{Statements and Declarations}

The authors declared that they have no conflicts of interest related to this work.

\bibliographystyle{bst/sn-basic}
\bibliography{ref}

\end{document}